\pdfoutput=1

\documentclass[11pt]{article}

\usepackage[final]{acl}

\usepackage{times}
\usepackage{latexsym}

\usepackage[T1]{fontenc}

\usepackage[utf8]{inputenc}

\usepackage{microtype}

\usepackage{inconsolata}
\usepackage{graphicx}
\usepackage{algorithm}
\usepackage{algorithmicx}
\usepackage{algpseudocode}
\usepackage{subcaption}
\usepackage{multirow}
\usepackage{rotating}
\usepackage{arydshln}
\usepackage{verbatim}
\usepackage{url}
\usepackage{amsmath}
\usepackage{amssymb}
\usepackage{mathtools}

\title{Don't Rank, Combine! Combining Machine Translation Hypotheses\\Using Quality Estimation}

\author{Giorgos Vernikos$^{1,2}$ \\[4pt] $^{1}$EPFL\\ Lausanne, Switzerland  \\
\hspace{7.5cm} \texttt{\{georgios.vernikos, andrei.popescu-belis\}@heig-vd.ch}
         \And
         Andrei Popescu-Belis$^{1,2}$ \\[4pt] $^{2}$HEIG-VD / HES-SO \\  Yverdon-les-Bains, Switzerland \\
}

\begin{document}
\maketitle
\begin{abstract}
Neural machine translation systems estimate probabilities of target sentences given source sentences, yet these estimates may not align with human preferences.  This work introduces QE-fusion, a method that synthesizes translations using a quality estimation metric (QE), which correlates better with human judgments. QE-fusion leverages a pool of candidates sampled from a model and combines spans from different candidates using a QE metric such as \textsc{CometKiwi}. We compare QE-fusion against beam search and recent reranking techniques, such as Minimum Bayes Risk decoding or QE-reranking. Our method consistently improves translation quality in terms of COMET and BLEURT scores when applied to large language models (LLMs) used for translation (PolyLM, XGLM, Llama2, Mistral, ALMA, and Tower) and to multilingual translation models (NLLB), over five language pairs. Notably, QE-fusion exhibits larger improvements for LLMs due to their ability to generate diverse outputs. 
We demonstrate that our approach generates novel translations in over half of the cases and consistently outperforms other methods across varying numbers of candidates (5–200). Furthermore, we empirically show that QE-fusion scales linearly with the number of candidates in the pool.
\end{abstract}

\section{Introduction}

Neural machine translation (NMT) models are probability estimators of translations given source sentences. Therefore, errors in NMT often arise due to the misalignment between the model's probabilities and human preferences~\cite{koehn-knowles-2017-six, Ott2018AnalyzingUI, stahlberg-byrne-2019-nmt}. In contrast, MT evaluation metrics have shown increasing correlation with human judgements~\cite{mathur-etal-2020-tangled, mqm, freitag-etal-2022-results}, making them valuable tools for selecting candidates generated by NMT models. 

Reference-based evaluation metrics such as COMET~\cite{rei-etal-2020-comet} and BLEURT~\cite{sellam-etal-2020-bleurt} have been successfully employed for Minimum Bayes Risk (MBR) decoding~\cite{fernandes-etal-2022-quality, freitag-etal-2022-high}. In MBR, these metrics serve as utility functions
to select the candidate with the highest similarity to all other candidates \cite{MBR, kumar-byrne-2004-minimum}. Additionally, quality estimation (QE) metrics such as COMET-QE~\cite{rei-etal-2021-references}, which do not need reference translations, have been used to rerank a pool of candidates based on their estimated quality \cite{fernandes-etal-2022-quality, farinhas-etal-2023-empirical}. Such reranking approaches improve translation quality over standard beam search, particularly when measured with neural-based MT evaluation metrics.

\begin{figure*}[ht]
    \centering
    \includegraphics[width=0.8\linewidth]{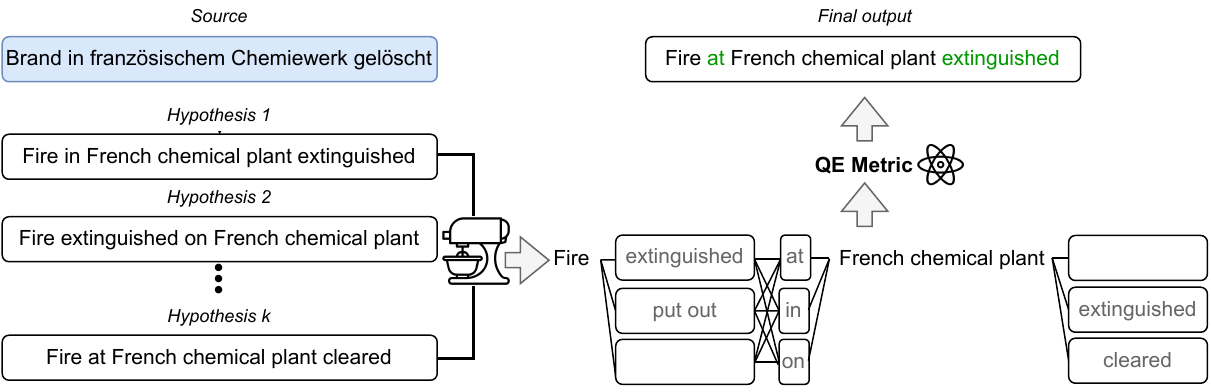}
  \caption{Illustration of the QE-fusion pipeline. The method first generates multiple hypotheses by sampling translations from the model. Then, it computes and sorts the spans that diverge among the candidates. Finally, a QE metric is used to select a span from each group and these spans are merged to form 
  a new, refined 
  translation.}
  \label{fig:approach}
\end{figure*}

However, reranking methods
are challenged in situations where candidate translations exhibit complementary errors, with no candidate clearly improving over all others. Figure~\ref{fig:approach} illustrates this issue.  The first candidate, \textit{Fire in French chemical plant}, uses \textit{in} instead of the more idiomatic \textit{at}. The third candidate makes a better choice in this case but renders the verb as \textit{cleared} instead of \textit{extinguished}, which is incorrect here.  Making the most of this pool of candidates is not possible without combining fragments from several ones.

To address this limitation, we propose QE-fusion, an approach that leverages the potential complementarity of model-generated candidates to synthesize an improved translation, guided by a QE metric.  QE-fusion starts by identifying divergent spans, i.e.\ spans that exhibit variations within the pool of candidates.  Then, it traverses these divergent spans, selecting at each step the one that contributes to a higher score according to the QE metric. The chosen spans are integrated into the synthesized translation, which is thus a fusion of multiple candidates.

We demonstrate that QE-fusion is effective with pools of candidates obtained from high-performing multilingual NMT models such as NLLB \cite{nllbteam2022language},
as well as with candidates obtained using in-context learning with large language models (LLMs), which have recently shown comparable performance to NMT models \cite{unreasonable_MT, hendy2023good, zhu2023multilingual}.  Indeed, QE-fusion improves translation for several popular, open-source LLMs:  PolyLM~\cite{wei2023polylm}, XGLM~\cite{lin-etal-2022-shot}, Llama2~\cite{touvron2023llama}, Mistral~\cite{jiang2023mistral}, ALMA~\cite{xu2023paradigm} and TowerBase~\cite{alves2024tower}.

Our contributions are the following:\footnote{The code is available at \url{https://github.com/GeorgeVern/qe-fusion}.}
\begin{enumerate}  \setlength{\itemsep}{0pt}
    \item We introduce QE-fusion, a novel algorithm that leverages quality estimation metrics to create improved translations by combining spans from a pool of candidates.
    \item We demonstrate the superior performance of QE-fusion compared to the recent QE-reranking and MBR decoding methods, across various open-source LLMs and multilingual NMT models, in five language pairs.
    \item We explain QE-fusion's larger improvements for LLMs over NMT models by the greater diversity of candidates generated by LLMs.
    \item We showcase the efficiency of our algorithm by 
    empirically demonstrating 
    that its runtime scales linearly with the number of candidates. 
    \item We show that QE-fusion effectively reduces hallucinations compared to QE-reranking and MBR, thus enhancing translation reliability.
\end{enumerate}

\section{Related Work}

We review in this section several approaches that aim to incorporate additional knowledge
either during the decoding process or after it, in order to mitigate the misalignment between MT models and human preferences.  We start with a brief reminder of MT evaluation metrics, which play a crucial role as a selection criterion.

\paragraph{MT evaluation metrics.} 
Traditional overlap-based evaluation metrics for MT such as BLEU~\cite{papineni-etal-2002-bleu} and ChrF~\cite{popovic-2015-chrf} are known for their imperfect correlation with human judgments \cite{mathur-etal-2020-tangled, kocmi-etal-2021-ship, mqm, freitag-etal-2022-results}. In response, researchers have shifted their attention towards metrics that use neural networks to score translation outputs, such as BERTScore \cite{Zhang*2020BERTScore:}, Prism \cite{thompson-post-2020-automatic}, COMET \cite{rei-etal-2020-comet} and BLEURT \cite{sellam-etal-2020-bleurt}. To better emulate human assessment, these metrics are often fine-tuned to predict human scores based on annotations such as those provided by the WMT metrics tasks~\cite{freitag-etal-2022-results}. 
In contrast to the reference-based metrics above, which have access to the reference translation, reference-free or \textit{quality estimation} metrics solely rely on the source sentence for estimating the quality of a translation \cite{zerva-etal-2022-findings}.  Recent QE metrics are COMET-QE \cite{rei-etal-2021-references}, TransQuest \cite{ranasinghe-etal-2020-transquest} and \textsc{CometKiwi} \cite{rei-etal-2022-cometkiwi}.

\paragraph{Reranking candidate translations.}
One of the earliest proposals for reranking the outputs of NMT systems is the adaptation of the noisy channel model~\cite{brown} to NMT systems~\cite{yee-etal-2019-simple, bhosale-etal-2020-language}. This approach combines scores from three components: a forward translation model trained on the original direction, a backward model trained on the opposite direction, and a language model trained on the target language. By integrating these diverse components, noisy-channel reranking has significantly improved translation quality \cite{ng-etal-2019-facebook}.

An alternative reranking approach uses Minimum Bayes Risk \cite{MBR, GOEL2000115}, which selects the candidate with the highest utility from a pool of candidates. 
In MBR, the utility function measures the similarity between a hypothesis and a reference; however, at test time, when references are unknown, the same set of hypotheses serves both as the candidates and the pseudo-references.  Possible utility functions include overlap-based \cite{eikema-aziz-2020-map} or state-of-the-art neural-based MT metrics \cite{eikema-aziz-2022-sampling, fernandes-etal-2022-quality, freitag-etal-2022-high}. MBR has shown promising results, 
particularly when equipped with recent automatic MT metrics. 
However, its time complexity scales quadratically with the size of the candidate pool and depends on the cost of computing the utility function, 
which is considerable for neural metrics such as COMET.

A more direct strategy is using reference-free metrics to rerank generated outputs, known as best-of-n or QE-reranking. Initial findings suggested that this approach resulted in inferior translations compared to beam search \cite{freitag-etal-2022-high, fernandes-etal-2022-quality}, especially due to the insensitivity of such metrics to hallucinations~\cite{guerreiro-etal-2023-looking}. 
However, with advancements in reference-free metrics, QE-reranking has 
outperformed beam search~\cite{gulcehre2023reinforced, finkelstein2023mbr}. 
Beyond reranking, QE scores can also be used during the training phase: for data filtering~\cite{finkelstein2023mbr}, curriculum construction~\cite{gulcehre2023reinforced}, or assigning quality-related tags to the outputs~\cite{tomani2023quality}. 
Notably, QE-reranking is substantially faster than MBR as its runtime scales linearly with the number of candidates.  
Yet, while reranking methods such as MBR decoding and QE-reranking effectively improve the performance of MT systems, they are ultimately constrained by the quality of the candidates in the pool. 

\paragraph{Post-editing MT outputs.}
An alternative to reranking is editing the actual outputs of the LLM or MT system using trained post-editing models~\cite{freitag-etal-2019-ape, voita-etal-2019-context, welleck2023generating}. While this approach can improve results, it requires training an additional model with a considerable amount of data and comparable size to the original model. Recent works have explored the complementarity of outputs by training models to merge multiple candidates generated by LLMs~\cite{jiang-etal-2023-llm, vernikos-etal-2024-small}. \citet{farinhas-etal-2023-empirical} used an LLM instead of a trained model to merge the candidates. Their experiments demonstrated comparable performance to reranking approaches that use neural MT metrics.

\paragraph{Constructing translation candidates.}
A wide variety of scores have been used to 
construct candidate translation,
including the model's own future scores~\cite{jinnai2023depth}, lookahead heuristics~\cite{lu-etal-2022-neurologic}, annd values predicted by separate models using Monte Carlo tree search \cite{leblond-etal-2021-machine, Liu2023DontTA, chaffin-etal-2022-ppl}. However, these approaches require access to the model, which is not always guaranteed, and entail substantial computational overhead due to the number of candidates stored at each time step. 


\section{Definition of QE-fusion} 
QE-fusion leverages the complementary nature of a pool of translation candidates, combining their strongest fragments into an improved output. These candidates are generated either by an LLM with an appropriate prompt or by an encoder-decoder NMT model.

\subsection{Candidate Generation}

There are multiple ways to generate candidates from a model. Common approaches for LLMs include nucleus or top-$p$ sampling~\cite{Holtzman2020The}, top-$k$, or sampling with a temperature. In our experiments, we adopt nucleus sampling with a temperature. The performance of LLMs can also be influenced by the number of samples and the prompt  \cite{bawden-yvon-2023-investigating, zhu2023multilingual}. To optimize LLM performance, we follow the guidelines of \citet{zhu2023multilingual} and use 8 examples for in-context learning in all our experiments.

For NMT models, beam search, while commonly used to return the top candidates in the beam, has been shown to lack diversity~\cite{vijayakumar2017diverse}. Instead, we use epsilon sampling~\cite{hewitt-etal-2022-truncation}, which sets the probability of tokens below a certain threshold to zero and was recently shown to generate more diverse candidates compared to beam search~\cite{freitag2023epsilon}. 

\subsection{QE-fusion Algorithm}


For each sentence translated by a model, the QE-fusion algorithm (Algorithm~\ref{alg:combine_hyps}) considers the top-ranked hypothesis by the QE metric as the base hypothesis, $h^{base}$ (line 1). The rationale behind this choice is that this is already a high-quality translation which may require the fewest modifications. The algorithm then identifies \emph{divergent spans} between $h^{base}$ and other candidates generated by the model, using an off-the-shelf library\footnote{\href{https://docs.python.org/3/library/difflib.html}{https://docs.python.org/3/library/difflib.html}} that employs edit distance to spot the additions, deletions and replacements required to transform one sequence into another (line 3). 

For each span where candidates diverge, 
the various alternatives are listed (line 4). 
The algorithm
substitutes the initial span of $h^{base}$ with each of the alternative spans (lines 6-11), synthesizing new hypotheses,
$\{h^{new}_i\}_{i=1,2,\ldots}$. A QE metric is used to compute scores for each $h^{new}_i$ (line 13). The hypotheses are then sorted based on their scores, and the top $b$ candidates are retained, forming a beam (lines 14-15). This process is repeated for each span of $h^{base}$, 
progressively creating new hypotheses. Finally, the highest-scoring hypothesis is selected. 

\renewcommand{\algorithmicrequire}{\textbf{Input:}}
\renewcommand{\algorithmicensure}{\textbf{Output:}}

\begin{algorithm} [t]
\small
\caption{QE-fusion algorithm} 
\label{alg:combine_hyps}
\begin{algorithmic}[1]
\Require candidate list $\mathcal{Y}$, QE metric $\mathcal{M}$, beam size $b$
\State $h^{base} \gets \underset{y \in \mathcal{Y}}{\mathrm{argmax}} \ \mathcal{M}(y)$ \Comment{select top-ranked candidate}
\State $hyps \gets \{h^{base}\}$ \Comment{initialize beam}
\State $\mathit{diffs} \gets$ \text{find\_diffs($h^{base}$, $\mathcal{Y}$)} \Comment{find divergent spans}
\For{$base\_span$, $alter\_spans$ \textbf{in} $\mathit{diffs}.items()$}
    \For{$h$ \textbf{in} $hyps$}
        \For{$span$ \textbf{in} $alter\_spans$}
            \State $h^{new} \gets h.\mathrm{replace}(base\_span, span)$ 
            \Statex \Comment{create new hypothesis}
            \If{$h^{new}$ \textbf{is not in} $hyps$}
                \State $hyps \gets hyps \cup \{h^{new}\}$
            \EndIf
        \EndFor
    \EndFor
    \State $scores \gets \mathcal{M}(hyps)$
    \State $sorted\_hyps \gets$ sorted($hyps$, $scores$) \Comment{sort hyps}
    \State $hyps \gets$ $sorted\_hyps[:b]$ \Comment{keep top $b$ hyps}
\EndFor
\Ensure $hyps[0]$
\end{algorithmic}
\end{algorithm}

While this is a conceptual explanation of the algorithm, certain modifications are made for efficiency purposes
such as batching the sentences and caching computed scores (see Appendix \ref{app:optimization}).

\section{Experimental Settings}

\subsection{Datasets and Evaluation Metrics}  

We assess performance across a spectrum of diverse language pairs. The selected languages include German, which is the most represented non-English language in most models; Russian, a high-resource language written in Cyrillic alphabet; Chinese, a high-resource language with a logographic script; and Icelandic, which is considered a low-resource language.  We consider the translation of English into the first two languages, and from the latter two into English.  Additionally, we consider a non-English-centric pair, German to French.

As we experiment with pre-trained models, we use only test data, which we take from WMT22 \cite{freitag-etal-2022-results} and for is$\rightarrow$en from WMT21 \cite{akhbardeh-etal-2021-findings}. We draw the few-shot examples for LLMs from the test sets of WMT21.
Further details regarding dataset sizes and domains are presented in the Appendix~\ref{app:data}.

We report results in terms of two neural-based metrics: COMET-22 \cite{rei-etal-2022-comet} and BLEURT-20 \cite{sellam-etal-2020-bleurt}. Surface-based BLEU and ChrF scores are given in Appendix~\ref{sec:bleu_results}. 

\subsection{Models and Parameters}

We conduct an extensive evaluation using LLMs and encoder-decoder NMT models of various sizes. Specifically, we use popular LLMs such as PolyLM-1.7B~\cite{wei2023polylm}, XGLM-2.9B~\cite{lin-etal-2022-shot}, Mistral-7B~\cite{jiang2023mistral}, Llama2-7B~\cite{touvron2023llama}. We additionally use ALMA-7B~\cite{xu2023paradigm} and TowerBase-7B~\cite{alves2024tower}, two LLMs based on Llama2-7B that are fine-tuned for translation using both monolingual and parallel data.
These LLMs have demonstrated impressive performance in MT~\cite{xu2023paradigm}, comparable to GPT-3.5 and NLLB-54B~\cite{nllbteam2022language}
and represent intermediary stages between general LLMs and task-specific MT models.
As for NMT systems, we use the multilingual NLLB-1.3B and 3.3B models~\cite{nllbteam2022language}. Results for the 13B versions of Llama2 and ALMA are given in Appendix~\ref{app:13b}. 

For the LLMs, we adopt the hyper-parameters used by \citet{touvron2023llama}. We generate candidates using nucleus sampling with $p=0.9$ and a temperature of 0.6. As a prompt for in-context learning, we follow \citet{zhu2023multilingual} and use the instruction template: $<$X$>$ = $<$Y$>$ with 8 examples randomly sampled from the WMT21 data. 
For the NMT models, we follow \citet{freitag2023epsilon} and use epsilon sampling \cite{hewitt-etal-2022-truncation} with $\epsilon=0.02$ and a temperature of 0.5. We study the impact of temperature in Section~\ref{sec:temp}. 
We generate 5 candidates both for LLM and MT models for efficiency purposes, although we demonstrate in Section~\ref{sec:pool_size} that our approach also works with larger numbers of candidates.


\begin{table*}[ht]
    \centering
    \small
    \begin{tabular}{lcccccc}
    \hline
        & \multicolumn{6}{c}{\bf LLM} \\
        \textbf{Method} & \textbf{PolyLM-1.7B} & \textbf{XGLM-2.9B}  &  \textbf{Llama2-7B} & \textbf{Mistral-7B} & \textbf{ALMA-7B} & \textbf{Tower-7B} \\ \hline
        & \multicolumn{6}{c}{en$\rightarrow$de}\\ 
        Greedy & 73.30 / 62.37 & 77.30 / 66.43 & 79.84 / 67.08 & 82.07 / 70.48 & 84.42 / 74.21 & 84.17 / 73.11\\
        Beam & 72.22 / 65.31 & 80.00 / 69.22 & 81.76 / 69.83 & 83.39 / 72.34 & 85.11 / 74.78 & 85.20 / 74.81\\ \hdashline
        Sample & 71.84 / 60.71 & 70.73 / 58.36 & 78.49 / 65.71 & 80.73 / 69.11 & 83.19 / 72.90 & 83.81 / 72.83\\ 
        MBR-BLEU & 73.54 / 62.08 & 77.18 / 65.77 & 79.34 / 66.73 & 81.72 / 69.99 & 84.04 / 73.73 & 84.35 / 73.25\\
        MBR-COMET & 78.68 / 66.36 & 80.90 / 68.80 & 83.26 / 70.32 & 84.73 / 72.78 & \textbf{85.99} / 75.46 & 86.17 / 74.63\\
        QE-reranking & 78.02 / 67.21 & 80.75 / 70.00 & 82.73 / 71.04 & 84.30 / 73.23 & 85.53 / 75.53 & 85.86 / 75.19\\ \hdashline
        QE-fusion & \textbf{79.62} / \textbf{68.67} & \textbf{81.62} / \textbf{71.01} & \textbf{83.63} / \textbf{71.96} & \textbf{85.02} / \textbf{74.10} & 85.93 / \textbf{75.93} & \textbf{86.23} / \textbf{75.68}\\ \hline
        & \multicolumn{6}{c}{en$\rightarrow$ru}\\ 
        Greedy & 74.52 / 59.70 & 80.74 / 65.75 & 81.72 / 66.94 & 84.85 / 71.13 & 85.80 / 72.31 & 86.94 / 74.24\\
        Beam & 74.79 / 63.26 & 82.11 / 67.58 & 83.72 / 69.49 & 86.32 / 73.13 & 86.87 / 74.17 & 87.34 / 75.07\\ \hdashline
        Sample & 72.99 / 57.70 & 71.65 / 52.45 & 80.18 / 64.90 & 83.55 / 69.17 & 84.84 / 70.81 & 86.36 / 73.23\\ 
        MBR-BLEU & 75.19 / 59.92 & 79.26 / 63.24 & 81.72 / 66.62 & 84.50 / 70.34 & 85.83 / 72.28 & 86.87 / 74.03\\
        MBR-COMET & 80.16 / 63.88 & 82.42 / 65.04 & 85.15 / 69.61 & 87.26 / 72.82 & 87.78 / 73.90 & \textbf{88.60} / 75.57\\
        QE-reranking & 79.59 / 64.53 & 83.07 / 68.07 & 84.62 / 70.11 & 86.81 / 73.25 & 87.33 / 74.32 & 88.29 / 75.90\\ \hdashline
        QE-fusion & \textbf{81.28} / \textbf{66.32} & \textbf{83.93} / \textbf{69.06} & \textbf{85.60} / \textbf{71.48} & \textbf{87.38} / \textbf{74.05} & \textbf{87.82} / \textbf{75.11} & 88.58 / \textbf{76.30}\\ \hline
        & \multicolumn{6}{c}{zh$\rightarrow$en}\\ 
        Greedy  & 68.69 / 54.76 & 46.51 / 28.76 & 77.72 / 64.13 & 79.69 / 66.67 & 79.24 / 66.20 & 78.49 / 65.30\\
        Beam & 66.61 / 57.71 & \textbf{73.94} / \textbf{58.69} & 78.52 / 65.89 & 80.08 / 67.74 & 79.23 / 66.46 & 78.49 / 65.95\\ \hdashline
        Sample & 67.85 / 53.32 & 53.77 / 33.70 & 76.55 / 62.69 & 78.91 / 65.56 & 78.19 / 64.70 & 77.73 / 64.03 \\ 
        MBR-BLEU & 69.74 / 54.83 & 62.93 / 44.64 & 77.80 / 64.12 & 79.69 / 66.58 & 79.09 / 65.81 & 78.33 / 64.80\\
        MBR-COMET & 72.47 / 56.49 & 65.98 / 46.51 & 79.30 / 65.05 & 80.87 / 67.31 & 80.40 / 66.77 & 79.79 / 66.01\\
        QE-reranking & 73.12 / 57.86 & 71.48 / 54.99 & 79.38 / 66.21 & 80.79 / 67.90 & 80.41 / 67.61 &  79.86 / 66.82\\ \hdashline
        QE-fusion & \textbf{74.27} / \textbf{59.06} & 72.14 / 55.80 & \textbf{79.99} / \textbf{66.92} & \textbf{81.15} / \textbf{68.44} & \textbf{80.86} / \textbf{68.13} & \textbf{80.44} / \textbf{67.46}\\ \hline
        & \multicolumn{6}{c}{de$\rightarrow$fr}\\ 
        Greedy & 63.09 / 41.88 & 71.50 / 53.40 & 76.88 / 60.46 & 78.65 / 63.16 & 74.86 / 57.48 & 80.64 / 65.86\\
        Beam & 62.35 / \textbf{50.45} & 74.32 / 57.37 & 78.61 / 63.69 & 79.66 / 65.50 & 77.07 / 60.66 & 81.30 / 67.52\\ \hdashline
        Sample & 61.27 / 38.59 & 67.62 / 45.94 & 75.24 / 57.60 & 77.12 / 60.60 & 72.81 / 53.85 & 79.55 / 64.50\\
        MBR-BLEU & 64.31 / 42.76 & 71.96 / 52.53 & 76.68 / 59.88 & 78.45 / 62.41 & 74.65 / 56.93 & 80.52 / 65.87\\
        MBR-COMET & 69.30 / 46.20 & 75.80 / 55.72 & 79.91 / 62.63 & 81.06 / 65.10 & 78.53 / 60.56 & \textbf{82.54} / 67.67\\
        QE-reranking & 68.74 / 48.38 & 75.51 / 57.31 & 79.44 / 63.67 & 80.69 / 65.53 & 77.96 / 61.18 & 82.16 / 68.07\\ \hdashline
        QE-fusion  & \textbf{70.09} / 49.98 & \textbf{76.64} / \textbf{58.76} & \textbf{80.27} / \textbf{64.56} & \textbf{81.26} / \textbf{66.28} & \textbf{78.87} / \textbf{62.48} & 82.53 / \textbf{68.44}\\ \hline
        & \multicolumn{6}{c}{is$\rightarrow$en}\\ 
        Greedy & -- & -- & 66.69 / 51.58 & 73.69 / 59.80 & 85.98 / 75.09 & 65.53 / 50.90\\
        Beam & -- & -- & 67.22 / 52.52 & 74.20 / 60.74 & 86.29 / 75.73 & 66.39 / 52.24\\ \hdashline
        Sample & -- & -- & 65.64 / 49.87 & 72.65 / 58.47 & 85.15 / 74.01 & 64.87 / 49.68\\
        MBR-BLEU & -- & -- & 66.33 / 50.96 & 73.24 / 59.22 & 85.79 / 74.81 & 65.74 / 50.80\\
        MBR-COMET & -- & -- & 69.32 / 52.36 & 75.57 / 60.81 & 86.58 / 75.46 & 68.77 / 52.28\\
        QE-reranking & -- & -- & 69.71 / 54.91 & 75.71 / 62.51 & 86.43 / 75.62 & 68.95 / 54.35\\ \hdashline
        QE-fusion  & -- & -- & \textbf{70.63} / \textbf{56.11} & \textbf{76.81} / \textbf{63.43} & \textbf{86.76} / \textbf{76.02} & \textbf{69.84} / \textbf{55.51}\\ \hline
    \end{tabular}
    \caption{Translation performance in terms of COMET-22~/ BLEURT-20 scores for various methods, language pairs, and sizes of LLMs. 
    Dashed lines separate deterministic decoding from sampling-based methods and from our approach.
    The best scores for each language pair and model are in \textbf{bold}.}
    \label{tab:llm_results}
\end{table*}

\begin{table}[ht]
    \centering
    \small
    \begin{tabular}{lcc}
        \hline
        & \multicolumn{2}{c}{\bf Multilingual NMT} \\
        \textbf{Method} &  \textbf{NLLB-1.3B} &  \textbf{NLLB-3.3B} \\ \hline
        & \multicolumn{2}{c}{en$\rightarrow$de}\\
        Greedy & 84.60 / 74.39 & 85.48 / 75.43 \\
        Beam & 85.54 / 75.62 & 86.24 / 76.44 \\ \hdashline
        Sample & 83.57 / 73.27 & 84.66 / 74.48 \\ 
        MBR-BLEU & 84.16 / 74.03 & 85.18 / 75.25 \\
        MBR-COMET & 85.98 / 75.38 & 86.69 / 76.34 \\
        QE-reranking & 85.92 / 75.88 & 86.25 / 76.60 \\ \hdashline
        QE-fusion & \textbf{86.25} / \textbf{76.11} & \textbf{86.74} / \textbf{76.81} \\ \hline
        & \multicolumn{2}{c}{en$\rightarrow$ru}\\
        Greedy  & 85.78 / 72.81 & 86.64 / 73.93 \\
        Beam  & 86.63 / 74.10 & 87.51 / 75.12 \\ \hdashline
        Sample  & 84.95 / 71.49 & 86.07 / 73.01 \\ 
        MBR-BLEU  & 85.59 / 72.37 & 86.44 / 73.58 \\
        MBR-COMET  & 87.47 / 73.97 & 88.18 / 75.09 \\
        QE-reranking  & 87.11 / 74.27 & 87.96 / 75.46 \\ \hdashline
        QE-fusion  & \textbf{87.52} / \textbf{74.71} & \textbf{88.31} / \textbf{75.83} \\ \hline
        & \multicolumn{2}{c}{zh$\rightarrow$en}\\ 
        Greedy & 72.38 / 59.35 & 73.64 / 60.81\\
        Beam & 75.16 / 62.90 & 75.88 / 63.97\\ \hdashline
        Sample & 71.39 / 57.45 & 72.20 / 58.50\\
        MBR-BLEU & 73.92 / 60.09  & 74.99 / 61.20\\
        MBR-COMET & 75.98 / 61.34 & 76.86 / 62.25\\
        QE-reranking & 76.46 / 62.84 & 77.41 / 63.90\\ \hdashline
        QE-fusion & \textbf{77.17} / \textbf{63.65} & \textbf{77.90} / \textbf{64.54}\\ \hline
        & \multicolumn{2}{c}{de$\rightarrow$fr}\\ 
        Greedy &  79.39 / 64.94 & 80.38 / 66.69 \\
        Beam & 81.03 / \textbf{67.69} & 81.92 / \textbf{69.24} \\ \hdashline
        Sample & 78.02 / 62.61 & 78.76 / 63.91 \\
        MBR-BLEU & 79.25 / 64.43 & 80.06 / 65.87 \\
        MBR-COMET & 81.43 / 66.20 & 82.15 / 67.63 \\
        QE-reranking & 81.21 / 66.86 & 81.88 / 68.26 \\ \hdashline
        QE-fusion & \textbf{81.82} / 67.63 & \textbf{82.33} / 68.81 \\ \hline
        & \multicolumn{2}{c}{is$\rightarrow$en}\\ 
        Greedy & 81.82 / 69.94 & 82.49 / 71.17 \\
        Beam & 82.30 / 70.60 & 83.05 / 71.71 \\ \hdashline
        Sample & 81.02 / 68.92 & 82.02 / 70.32 \\
        MBR-BLEU & 81.64 / 69.60 & 82.36 / 70.93 \\
        MBR-COMET & 82.74 / 70.39 & 83.56 / 71.71 \\
        QE-reranking & 82.81 / 71.04 & 83.50 / 71.97 \\ \hdashline
        QE-fusion & \textbf{83.08} / \textbf{71.45} & \textbf{83.83} / \textbf{72.33}\\ \hline
    \end{tabular}
    \caption{Translation performance in terms of COMET-22~/ BLEURT-20 scores for various methods, language pairs, and multilingual NMT models.}
    \label{tab:mmt_results}
\end{table}


\subsection{Comparison Terms} 

We compare QE-fusion with standard decoding algorithms for MT, such as greedy decoding and beam search with a width of 5. We also compare our approach with three sampling-based algorithms: (1)~random sampling from the pool, serving as a lower performance bound; (2)~QE-reranking~\cite{fernandes-etal-2022-quality} using \textsc{CometKiwi} \cite{rei-etal-2022-cometkiwi}, the same QE metric employed in QE-fusion; and (3)~MBR decoding~\cite{eikema-aziz-2020-map}, using either the surface-based BLEU or with the neural-based COMET-22 as the utility function.  As COMET-22 follows the same training pipeline and has the same number of parameters as \textsc{CometKiwi}, this ensures a fair comparison of the models. 

\section{Results: Translation Performance}

\subsection{QE-fusion Applied to LLMs} 

Table~\ref{tab:llm_results} presents the results obtained across various language pairs and LLMs, in terms of COMET and BLEURT scores.  Results with BLEU and ChrF scores, showing similar trends, are given in Appendix~\ref{sec:bleu_results}. As expected, the translation performance of LLMs generally improves with scale, but also with recency. For instance, the more recent Mistral-7B significantly outperforms Llama2-7B, despite their similar sizes (7 billion parameters). ALMA-7B and Tower-7B emerge as the top-performing LLMs across all language pairs, confirming the merits of MT-specific fine-tuning of LLMs.
Tower-7B has better performance than ALMA-7B in all pairs except is$\rightarrow$en, where the latter dominates.\footnote{This is because is$\rightarrow$en data was not used in the fine-tuning stages of Tower, contrary to ALMA, leading to catastrophic forgetting, as is visible when comparing the scores of Tower with those of its parent model, Llama2.}  We do not provide the is$\rightarrow$en scores of smaller LLMs (PolyLM and XGLM) due to their poor capabilities in the low-resource Icelandic language.

Regarding baselines, greedy decoding consistently lags behind beam search across all language pairs, an observation that contrasts with prior studies focused on zero-shot scenarios~\cite{farinhas-etal-2023-empirical}, likely due to our use of in-context examples. Unsurprisingly, random selection from the candidate pool emerges as the least effective baseline.

Among the reranking approaches, QE-reranking 
outperforms
MBR with either BLEU or COMET as the utility function, particularly in terms of BLEURT scores. Among the two utility functions for MBR, COMET is superior while the use of BLEU often fails to outperform beam search. MBR with COMET as the utility function occasionally surpasses QE-reranking in terms of COMET scores. This
may be due to a form of
``reward hacking'' \cite{gulcehre2023reinforced}, i.e.\ employing the same metric for both candidate selection and evaluation, since the BLEURT scores 
are in the reverse order.

Our approach, QE-fusion, consistently outperforms all
other methods,
across all language pairs and LLMs, with 5 exceptions out of 56 comparisons.  
Two notable exceptions are where beam search achieves the best BLEURT scores, for PolyLM-1.7B (de$\rightarrow$fr) and XGLM-2.9B (zh$\rightarrow$en).  In these cases, the candidate pool likely lacks high-quality translations altogether.  The other three exceptions correspond to very small differences in COMET scores (0.06, 0.02 and 0.01). 

Moreover, QE-fusion also outperforms the other methods when combined with even larger LLMs, as confirmed by the results obtained with the Llama2 and ALMA models with 13 billion parameters presented in Appendix~\ref{app:13b}, Tables~\ref{tab:extra_llm_results} and \ref{tab:extra_llm_results_bleu}, for COMET, BLEURT, BLEU and ChrF scores.

In Appendix~\ref{sec:radar-chart}, Figure~\ref{fig:spider_joint}, we present a graphical synthesis of these comparisons using radar charts.  The shapes corresponding to QE-fusion are always the outermost ones, regardless of variations due to the underlying LLM or language pair. 
In particular, our approach always outperforms QE-reranking, confirming its superiority as a generalization of reranking approaches.

\subsection{QE-fusion Applied to NMT Models} 

The COMET and BLEURT scores of our approach applied to 
multilingual NMT models, namely NLLB-1.3B and NLLB-3.3B, are presented in Table~\ref{tab:mmt_results} (BLEU and ChrF scores are in Appendix~\ref{sec:bleu_results}). We observe that NMT models have a slight edge over general-purpose LLMs (Llama2 and Mistral) while their MT-fine-tuned counterparts, ALMA and Tower, achieve similar or better performance.  Similar to the results with LLMs, our approach consistently outperforms beam search and reranking approaches.
The gap between our approach and QE-reranking is slightly smaller in the case of NMT models, which we attribute to the lower diversity of the generated candidates. We test this hypothesis in Section~\ref{sec:temp}.


\section{Analysis of Results} 

\subsection{Role of the Size of the Candidate Pool}
\label{sec:pool_size}

\begin{figure}[t]
    \centering
    \includegraphics[width=0.9\linewidth]{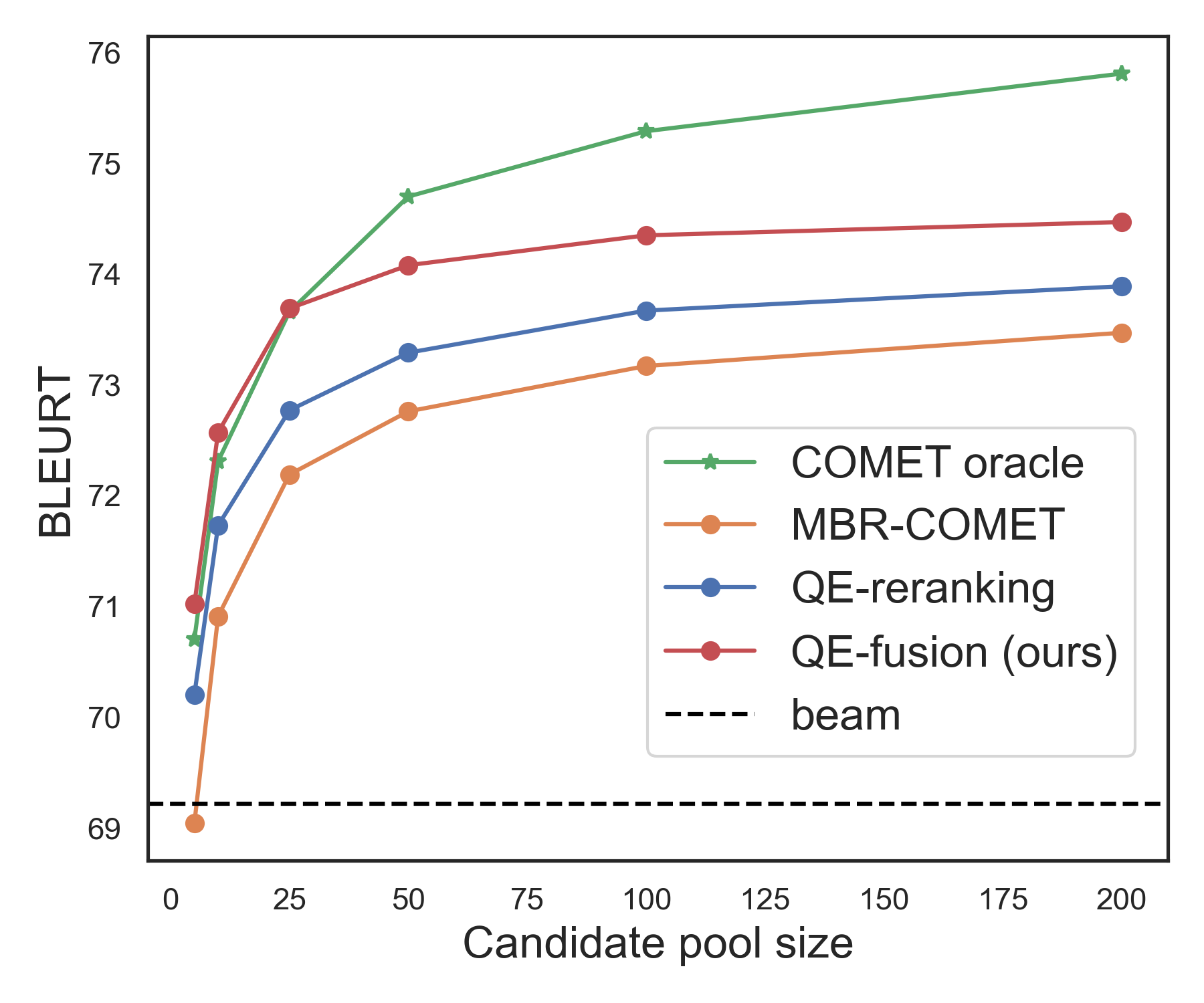}
    \caption{BLEURT scores of QE-fusion and other methods over pools of candidates of increasing sizes from the XGLM-2.9B LLM. QE-fusion outperforms reranking approaches and is comparable to the COMET-reranking \emph{oracle} for pools of up to 25 candidates.}
    \label{fig:pool_size}
\end{figure}

In the above experiments, we generated five candidate translations for efficiency reasons.  We now study the influence of the number of candidates on the scores of QE-fusion vs.\ those of the other methods, using XGLM-2.9B for en$\rightarrow$de translation.
We progressively sample larger candidate pools, from 5 to 200 candidates, and present in Figure~\ref{fig:pool_size} the BLEURT scores of our approach compared to QE-reranking using \textsc{CometKiwi} and to MBR using 
COMET
as the utility function. Additionally, we compare with an \emph{oracle} reranking approach that has access to the reference translation and uses COMET scores as the selection criterion. QE-fusion consistently outperforms reranking approaches across all sizes of candidate pools.  Moreover, QE-fusion even matches the performance of the oracle for pool sizes of 5, 10 and 25 candidates.

\subsection{Novelty of Outputs from QE-fusion}
\label{sec:novelty}

As the previous experiment may suggest that QE-fusion has a similar effect as the use of larger candidate pools with reranking methods, we examine here the novelty of the synthesized candidates, by counting how many times the output of QE-fusion can be found in a larger pool. For a pool of $p$ candidates given to QE-fusion, we measure how frequently an exact match of the output of QE-fusion can be found in larger pools of size $q \geq p$, where $p$ and $q$ are in \{5, 10, 25, 50, 100, 200\}. 

\begin{figure}[t]
    \includegraphics[width=0.9\linewidth]{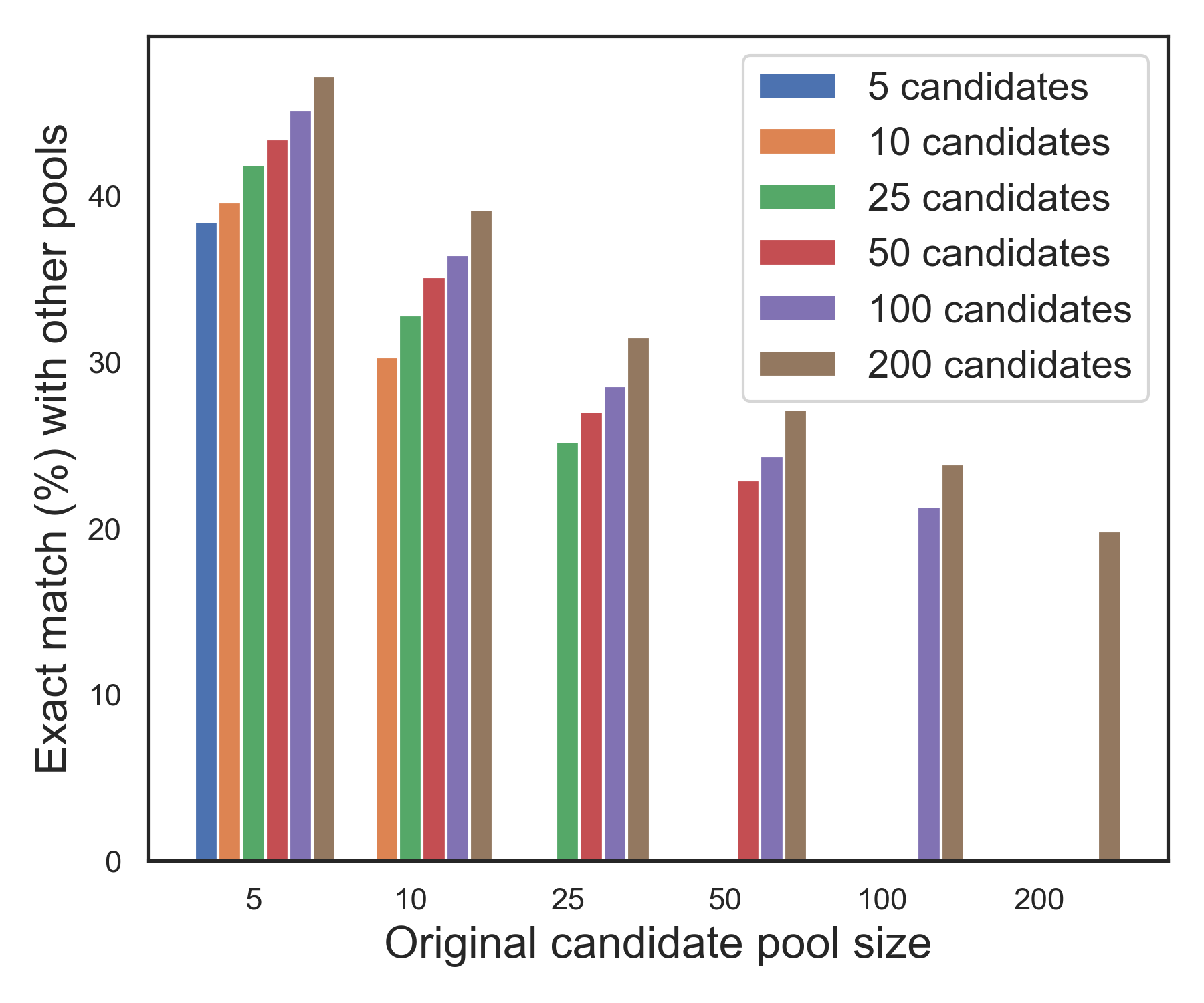} 
    \caption{Frequencies at which outputs produced by QE-fusion appear in larger candidate pools sampled from XGLM-2.9B.  Results show that QE-fusion always synthesizes a substantial number of novel candidates that the LLM would not generate otherwise.}
    \label{fig:novelty}
\end{figure}

The results, presented in Figure~\ref{fig:novelty}, reveal that even with a small pool of 5 candidates, more than 50\% of the outputs of QE-fusion would not have been generated by the LLM, even when sampling 200 candidates (rightmost bar of the leftmost group). The percentage of identical (or non-novel) candidates decreases as the pool grows, due to more varied candidates present in larger pools. 

When candidates generated by QE-fusion are present in the original pool, our method has the same effect as QE-reranking.  The frequency of these cases is given by the leftmost bar of each group (pool size) in Figure~\ref{fig:novelty}.  We observe that our approach defaults to QE-reranking less than 40\% of the time for a pool of size 5 (leftmost bar) and this value drops to 20\% for a pool of size 200.

\subsection{Impact of Candidate Diversity on Quality} 
\label{sec:temp}

By construction, QE-fusion benefits from the diversity of candidates, as this allows for an increased number of divergent spans. 
We study here the effect of diversity on both QE-fusion and QE-reranking, for LLMs and NMT models.

\begin{figure}
    \centering
    \includegraphics[width=0.85\linewidth]{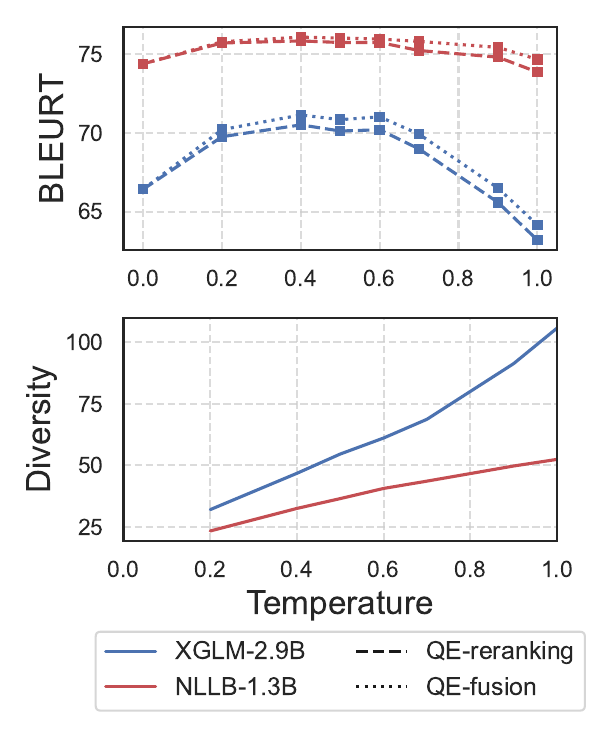}
    \caption{Effect of temperature on translation performance (above) and on the diversity of the pool (below), using an LLM and an NMT model for en$\rightarrow$de translation, with QE-fusion vs.\ QE-reranking.} 
    \label{fig:temp}
\end{figure}

To increase the diversity of the pool of candidate translations, we adjust the temperature parameter during decoding but keep constant all other generation parameters. 
A higher temperature results in token probability distributions that are more uniform, thus increasing the stochasticity of sampling and consequently the diversity of the candidate pool.  Here, we measure this diversity by the number of unique 4-grams present in the candidate pool, averaged over all test sentences. Figure~\ref{fig:temp} displays in its lower part the diversity of the pool as a function of temperature for XGLM-2.9B and NLLB-1.3B on en$\rightarrow$de translation, and in its upper part the BLEURT quality scores of these models with either QE-reranking or QE-fusion.  Additional results with COMET scores and other diversity measures are given in Appendix~\ref{app:temp-div} and show similar trends.

Increasing the temperature leads to an expected rise in diversity.  The rise is higher for XGLM-2.9B than for NLLB-1.3B, illustrating the fact that LLMs generate more diverse outputs, likely due to their general-domain language pretraining compared to the task-specific training of NMT models. Nevertheless, generating too diverse candidates due to high temperatures results in a noticeable drop in performance (right side of the upper graph).  The gap between QE-fusion and QE-reranking slightly widens as diversity increases, indicating the ability of our approach to leverage alternative spans. The optimal performance is achieved using a temperature in the [0.4, 0.6] interval. 

\subsection{Computation Time} \label{sec:runtime} 

In Section~\ref{sec:pool_size}, Figure~\ref{fig:pool_size}, we presented the scaling laws of QE-fusion vs.\ reranking in terms of performance when the size of the candidate pool varies. However, computation time is a crucial factor as the candidate pool grows. 
QE-reranking has the advantage of linear scaling with the number $N$ of candidates, while MBR requires $N(N-1)$ model calls for each sentence. In contrast, QE-fusion has variable complexity, depending on the diversity of the pool and the presence of different spans. 

\begin{figure}[t]
    \centering
    \includegraphics[width=0.9\linewidth]{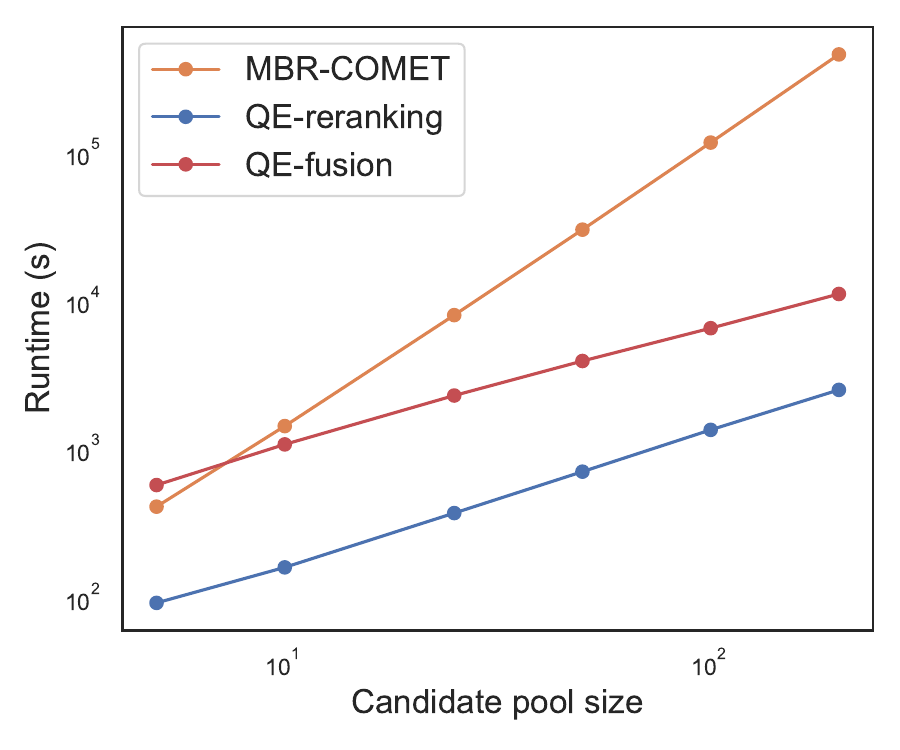}  
    \caption{Runtimes (in seconds) for different pool sizes for the en$\rightarrow$de WMT22 test set.} 
    \label{fig:runtime}
\end{figure}

To compare empirically the complexity of QE-fusion with reranking methods, we measured their runtime for the en$\rightarrow$de WMT22 test data with 2,037 sentences. All experiments were executed on a single Nvidia A40 GPU with 40 GB memory, using a batch size of 400 samples. Using a logarithmic scale, Figure~\ref{fig:runtime} confirms that QE-reranking scales linearly with the candidate pool size, while MBR scales quadratically. Interestingly, QE-fusion also exhibits linear scaling with the number of candidates but with a constant factor of $\times 5$ compared to QE-reranking.  For 5 and 10 candidates, QE-fusion has similar runtimes to MBR. 

We have implemented specific optimizations, including score caching and input batching, to reduce runtime (see Appendix \ref{app:optimization}). These modifications were uniformly applied to all methods to ensure a fair comparison. We leave further optimizations such as pruning~\cite{cheng-vlachos-2023-faster} for future work.

\subsection{Effect of QE-Fusion on Hallucinations}
Following the approach of \citet{guerreiro_hallucinations}, we define defect translations as those having an spBLEU score\footnote{We use spBLEU as implemented in SacreBLEU \cite{post-2018-call}: \texttt{nrefs:1|case:mixed|eff:yes|tok:flores101|\\smooth:exp|version:2.3.1}.} below 3, and compute their percentage. 
In our evaluation, unlike \citet{guerreiro_hallucinations}, we do not perturb the source sentences, but focus on measuring the frequency of hallucinations during the translation of the original unperturbed source sentences.

\begin{table}[t]
\centering
\resizebox{\linewidth}{!}{%
\setlength\tabcolsep{1.6pt}
\begin{tabular}{lccccc}
\hline
& \multicolumn{5}{c}{\bf Model} \\
\textbf{Method}	&\textbf{NLLB-1B}	&\textbf{NLLB-3B}	&\textbf{PolyLM} &\textbf{XGLM} &\textbf{Llama2} \\ \hline 
& \multicolumn{5}{c}{zh$\rightarrow$en}\\ 
Sample	& 15.6	& 15.9	& 20.4	& 59.4	& 5.8 \\
MBR-COMET & 9.3	& 9.0 & 16.3 & 35.4 & 4.1 \\
QE-reranking & 6.4 & \textbf{6.5} & 13.9 & 17.3 & 3.6 \\ \hdashline
QE-fusion & \textbf{5.5} & 6.7 & \textbf{12.5} & \textbf{16.1} & \textbf{3.1} \\ \hline
& \multicolumn{5}{c}{de$\rightarrow$fr}\\ 		
Sample & 3.6 & 3.8 & 27.4 & 20.4 & 5.5 \\
MBR-COMET & 2.4 & 2.4 & 21.6 & 9.3 & 3.6 \\
QE-reranking & 2.2 & 2.6 & 19.0 & 8.4 & 3.4 \\ \hdashline
QE-fusion & \textbf{2.1} & \textbf{2.2} & \textbf{17.7} & \textbf{7.7} & \textbf{3.0} \\ \hline
\end{tabular}
}
\caption{Percentage of defect translations (spBLEU $<$ 3) for various methods and models.}
\label{tab:hallucinations}
\end{table}

The results presented in Table~\ref{tab:hallucinations} indicate  that QE-reranking is more efficient in mitigating hallucinations compared to MBR with COMET, consistent with prior work \cite{farinhas-etal-2023-empirical}. Remarkably, QE-fusion reduces hallucinations even further, outperforming all other methods across different models and language pairs, in nine out of ten cases. 

\subsection{Comparison to Post-editing Approaches}

Due to the large scale and often restricted access of LLMs, novel approaches have emerged
to enhance their predictions without access to their weights. These methods, similar to reranking, involve generating predictions from the LLM and subsequently refining them using a secondary, more compact model  \cite{vernikos-etal-2024-small, jiang-etal-2023-llm}. 

We compare QE-fusion with LMCor~\cite{vernikos-etal-2024-small}, a T5-base \cite{t5} corrector model, which refines LLM-generated translations.
LMCor is trained on the News Commentary v16 corpus\footnote{\url{www.statmt.org/wmt22/translation-task.html}} augmented with five translation candidates from XGLM-2.9B for each source sentence. During testing, it receives the source sentence and five translation candidates. We also include results from a T5-base model fine-tuned for MT without additional candidates. 

\begin{table}[t]
\centering
\small
\begin{tabular}{lccc}
\hline
\textbf{Model} & \textbf{BLEU} & \textbf{COMET} & \textbf{BLEURT}\\ \hline
T5-base  (FT) & 23.32  &  75.22 & 64.57\\ \hline
XGLM-2.9B (ICL) & 17.32 & 74.54 & 66.47\\ \hdashline
+ \textsc{LMCor} & \textbf{25.15} & 77.45 & 68.41 \\ 
+ QE-fusion & 18.60 & \textbf{81.79} & \textbf{71.17}\\ \hline
\end{tabular}
\caption{Translation performance on the WMT22 en$\rightarrow$de test set. We compare QE-fusion against standard fine-tuning (FT), in-context learning (ICL), and  LMCor.}
\label{tab:mt_corrector}
\end{table}

Table~\ref{tab:mt_corrector} shows that LMCor outperforms fine-tuned T5 and in-context learning across all metrics. However, QE-fusion achieves significant improvements of 4 and 3 points in COMET and BLEURT, respectively, despite trailing in BLEU. This 
underscores the limitations of surface-based metrics and 
suggests that higher BLEU scores may be attained by models specifically trained for MT, which can mimic the translation style of the training data (see Appendix \ref{sec:bleu_results}). 

\section{Conclusion}

In this paper, we introduced QE-fusion, a novel approach that combines fragments from diverse candidates to synthesize improved translations guided by quality estimation metrics. Our experiments across five language pairs involving both LLMs and NMT models demonstrate that QE-fusion consistently outperforms beam search, MBR decoding, and QE-reranking. QE-fusion is particularly beneficial to LLMs, capitalizing on their diverse outputs. Notably, QE-fusion maintains its superior performance even when the pool of candidates grows, while its runtime scales linearly with the number of candidates, highlighting its scalability. 

By integrating external knowledge from quality estimation metrics, QE-fusion not only enhances translation quality but also tackles specific issues like hallucinations. This offers a promising avenue for addressing other phenomena such as context-specific errors~\cite{vernikos-etal-2022-embarrassingly}. Beyond translation, the versatility of QE-fusion extends to any text generation task where metrics or reward models \cite{ouyang2022training} are available.

\section*{Limitations}
\paragraph{Human evaluation.} While our work employs state-of-the-art MT evaluation metrics, we acknowledge the inherent limitations of automatic metrics. Human evaluation could offer additional, and possibly more reliable insights. At the time of writing, human evaluation is ongoing, starting with one language pair and four systems.

\paragraph{Choice of metrics.} 
The QE metric used by QE-fusion, \textsc{COMETKiwi}, shares some similarities with the COMET metric used for evaluation, as they originate from the same family of models. Consequently, using \textsc{COMETKiwi} as our criterion for merging spans might be considered as the reason why we get improvements in COMET scores. To address this concern, we confirm our findings with scores using three other metrics: BLEURT, BLEU and ChrF.  Even with these alternate metrics, our approach consistently outperforms all other reranking techniques.

\section*{Acknowledgments}
We are grateful for their support to the Swiss National Science Foundation (DOMAT grant n.\ 175693, On-demand Knowledge for Document-level Machine Translation and EXOMAT grant n.\ 228494, External Knowledge for Low-resource Machine Translation), and to the Institute for ICT at HEIG-VD.

\bibliography{anthology,custom}

\begin{thebibliography}{67}
\expandafter\ifx\csname natexlab\endcsname\relax\def\natexlab#1{#1}\fi

\bibitem[{Akhbardeh et~al.(2021)Akhbardeh, Arkhangorodsky, Biesialska, Bojar, Chatterjee, Chaudhary, Costa-jussa, Espa{\~n}a-Bonet, Fan, Federmann, Freitag, Graham, Grundkiewicz, Haddow, Harter, Heafield, Homan, Huck, Amponsah-Kaakyire, Kasai, Khashabi, Knight, Kocmi, Koehn, Lourie, Monz, Morishita, Nagata, Nagesh, Nakazawa, Negri, Pal, Tapo, Turchi, Vydrin, and Zampieri}]{akhbardeh-etal-2021-findings}
Farhad Akhbardeh, Arkady Arkhangorodsky, Magdalena Biesialska, Ond{\v{r}}ej Bojar, Rajen Chatterjee, Vishrav Chaudhary, Marta~R. Costa-jussa, Cristina Espa{\~n}a-Bonet, Angela Fan, Christian Federmann, Markus Freitag, Yvette Graham, Roman Grundkiewicz, Barry Haddow, Leonie Harter, Kenneth Heafield, Christopher Homan, Matthias Huck, Kwabena Amponsah-Kaakyire, Jungo Kasai, Daniel Khashabi, Kevin Knight, Tom Kocmi, Philipp Koehn, Nicholas Lourie, Christof Monz, Makoto Morishita, Masaaki Nagata, Ajay Nagesh, Toshiaki Nakazawa, Matteo Negri, Santanu Pal, Allahsera~Auguste Tapo, Marco Turchi, Valentin Vydrin, and Marcos Zampieri. 2021.
\newblock \href {https://aclanthology.org/2021.wmt-1.1} {Findings of the 2021 conference on machine translation ({WMT}21)}.
\newblock In \emph{Proceedings of the Sixth Conference on Machine Translation}, pages 1--88, Online. Association for Computational Linguistics.

\bibitem[{Alves et~al.(2024)Alves, Pombal, Guerreiro, Martins, Alves, Farajian, Peters, Rei, Fernandes, Agrawal, Colombo, de~Souza, and Martins}]{alves2024tower}
Duarte~M. Alves, José Pombal, Nuno~M. Guerreiro, Pedro~H. Martins, João Alves, Amin Farajian, Ben Peters, Ricardo Rei, Patrick Fernandes, Sweta Agrawal, Pierre Colombo, José G.~C. de~Souza, and André F.~T. Martins. 2024.
\newblock \href {https://arxiv.org/abs/2402.17733} {Tower: An open multilingual large language model for translation-related tasks}.
\newblock \emph{arXiv preprint arXiv:2402.17733}.

\bibitem[{Bawden and Yvon(2023)}]{bawden-yvon-2023-investigating}
Rachel Bawden and Fran{\c{c}}ois Yvon. 2023.
\newblock \href {https://aclanthology.org/2023.eamt-1.16} {Investigating the translation performance of a large multilingual language model: the case of {BLOOM}}.
\newblock In \emph{Proceedings of the 24th Annual Conference of the European Association for Machine Translation}, pages 157--170, Tampere, Finland. European Association for Machine Translation.

\bibitem[{Bhosale et~al.(2020)Bhosale, Yee, Edunov, and Auli}]{bhosale-etal-2020-language}
Shruti Bhosale, Kyra Yee, Sergey Edunov, and Michael Auli. 2020.
\newblock \href {https://aclanthology.org/2020.wmt-1.69} {Language models not just for pre-training: Fast online neural noisy channel modeling}.
\newblock In \emph{Proceedings of the Fifth Conference on Machine Translation}, pages 584--593, Online. Association for Computational Linguistics.

\bibitem[{Brown et~al.(1993)Brown, Della~Pietra, Della~Pietra, and Mercer}]{brown}
Peter~F. Brown, Stephen~A. Della~Pietra, Vincent~J. Della~Pietra, and Robert~L. Mercer. 1993.
\newblock \href {https://aclanthology.org/J93-2003} {The mathematics of statistical machine translation: Parameter estimation}.
\newblock \emph{Computational Linguistics}, 19(2):263--311.

\bibitem[{Chaffin et~al.(2022)Chaffin, Claveau, and Kijak}]{chaffin-etal-2022-ppl}
Antoine Chaffin, Vincent Claveau, and Ewa Kijak. 2022.
\newblock \href {https://doi.org/10.18653/v1/2022.naacl-main.215} {{PPL-MCTS}: {C}onstrained textual generation through discriminator-guided {MCTS} decoding}.
\newblock In \emph{Proceedings of the 2022 Conference of the North American Chapter of the Association for Computational Linguistics: Human Language Technologies}, pages 2953--2967, Seattle, United States. Association for Computational Linguistics.

\bibitem[{Cheng and Vlachos(2023)}]{cheng-vlachos-2023-faster}
Julius Cheng and Andreas Vlachos. 2023.
\newblock \href {https://doi.org/10.18653/v1/2023.emnlp-main.767} {Faster minimum {B}ayes risk decoding with confidence-based pruning}.
\newblock In \emph{Proceedings of the 2023 Conference on Empirical Methods in Natural Language Processing}, pages 12473--12480.

\bibitem[{Chowdhery et~al.(2023)Chowdhery, Narang, Devlin, Bosma, Mishra, Roberts, Barham, Chung, Sutton, Gehrmann, Schuh, Shi, Tsvyashchenko, Maynez, Rao, Barnes, Tay, Shazeer, Prabhakaran, Reif, Du, Hutchinson, Pope, Bradbury, Austin, Isard, Gur-Ari, Yin, Duke, Levskaya, Ghemawat, Dev, Michalewski, Garcia, Misra, Robinson, Fedus, Zhou, Ippolito, Luan, Lim, Zoph, Spiridonov, Sepassi, Dohan, Agrawal, Omernick, Dai, Pillai, Pellat, Lewkowycz, Moreira, Child, Polozov, Lee, Zhou, Wang, Saeta, Diaz, Firat, Catasta, Wei, Meier-Hellstern, Eck, Dean, Petrov, and Fiedel}]{palm}
Aakanksha Chowdhery, Sharan Narang, Jacob Devlin, Maarten Bosma, Gaurav Mishra, Adam Roberts, Paul Barham, Hyung~Won Chung, Charles Sutton, Sebastian Gehrmann, Parker Schuh, Kensen Shi, Sasha Tsvyashchenko, Joshua Maynez, Abhishek Rao, Parker Barnes, Yi~Tay, Noam Shazeer, Vinodkumar Prabhakaran, Emily Reif, Nan Du, Ben Hutchinson, Reiner Pope, James Bradbury, Jacob Austin, Michael Isard, Guy Gur-Ari, Pengcheng Yin, Toju Duke, Anselm Levskaya, Sanjay Ghemawat, Sunipa Dev, Henryk Michalewski, Xavier Garcia, Vedant Misra, Kevin Robinson, Liam Fedus, Denny Zhou, Daphne Ippolito, David Luan, Hyeontaek Lim, Barret Zoph, Alexander Spiridonov, Ryan Sepassi, David Dohan, Shivani Agrawal, Mark Omernick, Andrew~M. Dai, Thanumalayan~Sankaranarayana Pillai, Marie Pellat, Aitor Lewkowycz, Erica Moreira, Rewon Child, Oleksandr Polozov, Katherine Lee, Zongwei Zhou, Xuezhi Wang, Brennan Saeta, Mark Diaz, Orhan Firat, Michele Catasta, Jason Wei, Kathy Meier-Hellstern, Douglas Eck, Jeff Dean, Slav Petrov, and Noah Fiedel. 2023.
\newblock \href {http://jmlr.org/papers/v24/22-1144.html} {Palm: Scaling language modeling with pathways}.
\newblock \emph{Journal of Machine Learning Research}, 24(240):1--113.

\bibitem[{Eikema and Aziz(2020)}]{eikema-aziz-2020-map}
Bryan Eikema and Wilker Aziz. 2020.
\newblock \href {https://doi.org/10.18653/v1/2020.coling-main.398} {Is {MAP} decoding all you need? the inadequacy of the mode in neural machine translation}.
\newblock In \emph{Proceedings of the 28th International Conference on Computational Linguistics}, pages 4506--4520, Barcelona, Spain (Online). International Committee on Computational Linguistics.

\bibitem[{Eikema and Aziz(2022)}]{eikema-aziz-2022-sampling}
Bryan Eikema and Wilker Aziz. 2022.
\newblock \href {https://doi.org/10.18653/v1/2022.emnlp-main.754} {Sampling-based approximations to minimum {B}ayes risk decoding for neural machine translation}.
\newblock In \emph{Proceedings of the 2022 Conference on Empirical Methods in Natural Language Processing}, pages 10978--10993, Abu Dhabi, United Arab Emirates. Association for Computational Linguistics.

\bibitem[{Farinhas et~al.(2023)Farinhas, de~Souza, and Martins}]{farinhas-etal-2023-empirical}
Ant{\'o}nio Farinhas, Jos{\'e} de~Souza, and Andre Martins. 2023.
\newblock \href {https://doi.org/10.18653/v1/2023.emnlp-main.733} {An empirical study of translation hypothesis ensembling with large language models}.
\newblock In \emph{Proceedings of the 2023 Conference on Empirical Methods in Natural Language Processing}, pages 11956--11970, Singapore.

\bibitem[{Fernandes et~al.(2022)Fernandes, Farinhas, Rei, C.~de Souza, Ogayo, Neubig, and Martins}]{fernandes-etal-2022-quality}
Patrick Fernandes, Ant{\'o}nio Farinhas, Ricardo Rei, Jos{\'e}~G. C.~de Souza, Perez Ogayo, Graham Neubig, and Andre Martins. 2022.
\newblock \href {https://doi.org/10.18653/v1/2022.naacl-main.100} {Quality-aware decoding for neural machine translation}.
\newblock In \emph{Proceedings of the 2022 Conference of the North American Chapter of the Association for Computational Linguistics: Human Language Technologies}, pages 1396--1412, Seattle, United States. Association for Computational Linguistics.

\bibitem[{Finkelstein et~al.(2023)Finkelstein, Naskar, Mirzazadeh, Shah, and Freitag}]{finkelstein2023mbr}
Mara Finkelstein, Subhajit Naskar, Mehdi Mirzazadeh, Apurva Shah, and Markus Freitag. 2023.
\newblock \href {https://arxiv.org/abs/2309.10966} {{MBR} and {QE} finetuning: Training-time distillation of the best and most expensive decoding methods}.
\newblock \emph{arXiv preprint arXiv:2309.10966}.

\bibitem[{Freitag et~al.(2019)Freitag, Caswell, and Roy}]{freitag-etal-2019-ape}
Markus Freitag, Isaac Caswell, and Scott Roy. 2019.
\newblock \href {https://doi.org/10.18653/v1/W19-5204} {{APE} at scale and its implications on {MT} evaluation biases}.
\newblock In \emph{Proceedings of the Fourth Conference on Machine Translation (Volume 1: Research Papers)}, pages 34--44, Florence, Italy. Association for Computational Linguistics.

\bibitem[{Freitag et~al.(2021)Freitag, Foster, Grangier, Ratnakar, Tan, and Macherey}]{mqm}
Markus Freitag, George Foster, David Grangier, Viresh Ratnakar, Qijun Tan, and Wolfgang Macherey. 2021.
\newblock \href {https://doi.org/10.1162/tacl_a_00437} {{Experts, Errors, and Context: A Large-Scale Study of Human Evaluation for Machine Translation}}.
\newblock \emph{Transactions of the Association for Computational Linguistics}, 9:1460--1474.

\bibitem[{Freitag et~al.(2023)Freitag, Ghorbani, and Fernandes}]{freitag2023epsilon}
Markus Freitag, Behrooz Ghorbani, and Patrick Fernandes. 2023.
\newblock \href {https://aclanthology.org/2023.findings-emnlp.617} {Epsilon sampling rocks: Investigating sampling strategies for minimum {B}ayes risk decoding for machine translation}.
\newblock In \emph{Findings of the Association for Computational Linguistics: EMNLP 2023}, pages 9198--9209.

\bibitem[{Freitag et~al.(2022{\natexlab{a}})Freitag, Grangier, Tan, and Liang}]{freitag-etal-2022-high}
Markus Freitag, David Grangier, Qijun Tan, and Bowen Liang. 2022{\natexlab{a}}.
\newblock \href {https://doi.org/10.1162/tacl_a_00491} {High quality rather than high model probability: Minimum {B}ayes risk decoding with neural metrics}.
\newblock \emph{Transactions of the Association for Computational Linguistics}, 10:811--825.

\bibitem[{Freitag et~al.(2022{\natexlab{b}})Freitag, Rei, Mathur, Lo, Stewart, Avramidis, Kocmi, Foster, Lavie, and Martins}]{freitag-etal-2022-results}
Markus Freitag, Ricardo Rei, Nitika Mathur, Chi-kiu Lo, Craig Stewart, Eleftherios Avramidis, Tom Kocmi, George Foster, Alon Lavie, and Andr{\'e} F.~T. Martins. 2022{\natexlab{b}}.
\newblock \href {https://aclanthology.org/2022.wmt-1.2} {Results of {WMT}22 metrics shared task: Stop using {BLEU} {--} neural metrics are better and more robust}.
\newblock In \emph{Proceedings of the Seventh Conference on Machine Translation (WMT)}, pages 46--68, Abu Dhabi, United Arab Emirates (Hybrid). Association for Computational Linguistics.

\bibitem[{Garcia et~al.(2023)Garcia, Bansal, Cherry, Foster, Krikun, Johnson, and Firat}]{unreasonable_MT}
Xavier Garcia, Yamini Bansal, Colin Cherry, George Foster, Maxim Krikun, Melvin Johnson, and Orhan Firat. 2023.
\newblock \href {https://dl.acm.org/doi/10.5555/3618408.3618846} {The unreasonable effectiveness of few-shot learning for machine translation}.
\newblock In \emph{Proceedings of the 40th International Conference on Machine Learning}.

\bibitem[{Goel and Byrne(2000)}]{GOEL2000115}
Vaibhava Goel and William~J Byrne. 2000.
\newblock \href {https://doi.org/https://doi.org/10.1006/csla.2000.0138} {Minimum {B}ayes-risk automatic speech recognition}.
\newblock \emph{Computer Speech \& Language}, 14(2):115--135.

\bibitem[{Guerreiro et~al.(2023{\natexlab{a}})Guerreiro, Alves, Waldendorf, Haddow, Birch, Colombo, and Martins}]{guerreiro_hallucinations}
Nuno~M. Guerreiro, Duarte~M. Alves, Jonas Waldendorf, Barry Haddow, Alexandra Birch, Pierre Colombo, and André F.~T. Martins. 2023{\natexlab{a}}.
\newblock \href {https://doi.org/10.1162/tacl_a_00615} {{Hallucinations in Large Multilingual Translation Models}}.
\newblock \emph{Transactions of the Association for Computational Linguistics}, 11:1500--1517.

\bibitem[{Guerreiro et~al.(2023{\natexlab{b}})Guerreiro, Voita, and Martins}]{guerreiro-etal-2023-looking}
Nuno~M. Guerreiro, Elena Voita, and Andr{\'e} Martins. 2023{\natexlab{b}}.
\newblock \href {https://doi.org/10.18653/v1/2023.eacl-main.75} {Looking for a needle in a haystack: A comprehensive study of hallucinations in neural machine translation}.
\newblock In \emph{Proceedings of the 17th Conference of the European Chapter of the Association for Computational Linguistics}, pages 1059--1075, Dubrovnik, Croatia. Association for Computational Linguistics.

\bibitem[{Gulcehre et~al.(2023)Gulcehre, Paine, Srinivasan, Konyushkova, Weerts, Sharma, Siddhant, Ahern, Wang, Gu, Macherey, Doucet, Firat, and de~Freitas}]{gulcehre2023reinforced}
Caglar Gulcehre, Tom~Le Paine, Srivatsan Srinivasan, Ksenia Konyushkova, Lotte Weerts, Abhishek Sharma, Aditya Siddhant, Alex Ahern, Miaosen Wang, Chenjie Gu, Wolfgang Macherey, Arnaud Doucet, Orhan Firat, and Nando de~Freitas. 2023.
\newblock \href {https://arxiv.org/abs/2308.08998} {Reinforced self-training {(ReST)} for language modeling}.
\newblock \emph{arXiv preprint arXiv:2308.08998}.

\bibitem[{Hendy et~al.(2023)Hendy, Abdelrehim, Sharaf, Raunak, Gabr, Matsushita, Kim, Afify, and Awadalla}]{hendy2023good}
Amr Hendy, Mohamed Abdelrehim, Amr Sharaf, Vikas Raunak, Mohamed Gabr, Hitokazu Matsushita, Young~Jin Kim, Mohamed Afify, and Hany~Hassan Awadalla. 2023.
\newblock \href {https://arxiv.org/abs/2302.09210} {How good are {GPT} models at machine translation? a comprehensive evaluation}.
\newblock \emph{arXiv preprint arXiv:2302.09210}.

\bibitem[{Hewitt et~al.(2022)Hewitt, Manning, and Liang}]{hewitt-etal-2022-truncation}
John Hewitt, Christopher Manning, and Percy Liang. 2022.
\newblock \href {https://doi.org/10.18653/v1/2022.findings-emnlp.249} {Truncation sampling as language model desmoothing}.
\newblock In \emph{Findings of the Association for Computational Linguistics: EMNLP 2022}, pages 3414--3427, Abu Dhabi, United Arab Emirates. Association for Computational Linguistics.

\bibitem[{Holtzman et~al.(2020)Holtzman, Buys, Du, Forbes, and Choi}]{Holtzman2020The}
Ari Holtzman, Jan Buys, Li~Du, Maxwell Forbes, and Yejin Choi. 2020.
\newblock \href {https://openreview.net/forum?id=rygGQyrFvH} {The curious case of neural text degeneration}.
\newblock In \emph{The Eighth International Conference on Learning Representations}.

\bibitem[{Jiang et~al.(2023{\natexlab{a}})Jiang, Sablayrolles, Mensch, Bamford, Chaplot, de~las Casas, Bressand, Lengyel, Lample, Saulnier, Lavaud, Lachaux, Stock, Scao, Lavril, Wang, Lacroix, and Sayed}]{jiang2023mistral}
Albert~Q. Jiang, Alexandre Sablayrolles, Arthur Mensch, Chris Bamford, Devendra~Singh Chaplot, Diego de~las Casas, Florian Bressand, Gianna Lengyel, Guillaume Lample, Lucile Saulnier, Lélio~Renard Lavaud, Marie-Anne Lachaux, Pierre Stock, Teven~Le Scao, Thibaut Lavril, Thomas Wang, Timothée Lacroix, and William~El Sayed. 2023{\natexlab{a}}.
\newblock \href {https://arxiv.org/abs/2310.06825} {Mistral {7B}}.
\newblock \emph{arXiv preprint arXiv:2310.06825}.

\bibitem[{Jiang et~al.(2023{\natexlab{b}})Jiang, Ren, and Lin}]{jiang-etal-2023-llm}
Dongfu Jiang, Xiang Ren, and Bill~Yuchen Lin. 2023{\natexlab{b}}.
\newblock \href {https://doi.org/10.18653/v1/2023.acl-long.792} {{LLM}-blender: Ensembling large language models with pairwise ranking and generative fusion}.
\newblock In \emph{Proceedings of the 61st Annual Meeting of the Association for Computational Linguistics (Volume 1: Long Papers)}, pages 14165--14178, Toronto, Canada. Association for Computational Linguistics.

\bibitem[{Jinnai et~al.(2023)Jinnai, Morimura, and Honda}]{jinnai2023depth}
Yuu Jinnai, Tetsuro Morimura, and Ukyo Honda. 2023.
\newblock \href {https://arxiv.org/abs/2308.13696} {On the depth between beam search and exhaustive search for text generation}.
\newblock \emph{arXiv preprint arXiv:2308.13696}.

\bibitem[{Kocmi et~al.(2021)Kocmi, Federmann, Grundkiewicz, Junczys-Dowmunt, Matsushita, and Menezes}]{kocmi-etal-2021-ship}
Tom Kocmi, Christian Federmann, Roman Grundkiewicz, Marcin Junczys-Dowmunt, Hitokazu Matsushita, and Arul Menezes. 2021.
\newblock \href {https://aclanthology.org/2021.wmt-1.57} {To ship or not to ship: An extensive evaluation of automatic metrics for machine translation}.
\newblock In \emph{Proceedings of the Sixth Conference on Machine Translation}, pages 478--494, Online. Association for Computational Linguistics.

\bibitem[{Koehn and Knowles(2017)}]{koehn-knowles-2017-six}
Philipp Koehn and Rebecca Knowles. 2017.
\newblock \href {https://doi.org/10.18653/v1/W17-3204} {Six challenges for neural machine translation}.
\newblock In \emph{Proceedings of the First Workshop on Neural Machine Translation}, pages 28--39, Vancouver. Association for Computational Linguistics.

\bibitem[{Kumar and Byrne(2002)}]{MBR}
Shankar Kumar and William Byrne. 2002.
\newblock \href {https://doi.org/10.3115/1118693.1118712} {Minimum {B}ayes-risk word alignments of bilingual texts}.
\newblock In \emph{Proceedings of the 2002 Conference on Empirical Methods in Natural Language Processing}, page 140–147.

\bibitem[{Kumar and Byrne(2004)}]{kumar-byrne-2004-minimum}
Shankar Kumar and William Byrne. 2004.
\newblock \href {https://aclanthology.org/N04-1022} {Minimum {B}ayes-risk decoding for statistical machine translation}.
\newblock In \emph{Proceedings of the Human Language Technology Conference of the North {A}merican Chapter of the Association for Computational Linguistics: {HLT}-{NAACL} 2004}, pages 169--176, Boston, Massachusetts, USA. Association for Computational Linguistics.

\bibitem[{Leblond et~al.(2021)Leblond, Alayrac, Sifre, Pislar, Jean-Baptiste, Antonoglou, Simonyan, and Vinyals}]{leblond-etal-2021-machine}
R{\'e}mi Leblond, Jean-Baptiste Alayrac, Laurent Sifre, Miruna Pislar, Lespiau Jean-Baptiste, Ioannis Antonoglou, Karen Simonyan, and Oriol Vinyals. 2021.
\newblock \href {https://doi.org/10.18653/v1/2021.emnlp-main.662} {Machine translation decoding beyond beam search}.
\newblock In \emph{Proceedings of the 2021 Conference on Empirical Methods in Natural Language Processing}, pages 8410--8434, Online and Punta Cana, Dominican Republic. Association for Computational Linguistics.

\bibitem[{Lin et~al.(2022)Lin, Mihaylov, Artetxe, Wang, Chen, Simig, Ott, Goyal, Bhosale, Du, Pasunuru, Shleifer, Koura, Chaudhary, O{'}Horo, Wang, Zettlemoyer, Kozareva, Diab, Stoyanov, and Li}]{lin-etal-2022-shot}
Xi~Victoria Lin, Todor Mihaylov, Mikel Artetxe, Tianlu Wang, Shuohui Chen, Daniel Simig, Myle Ott, Naman Goyal, Shruti Bhosale, Jingfei Du, Ramakanth Pasunuru, Sam Shleifer, Punit~Singh Koura, Vishrav Chaudhary, Brian O{'}Horo, Jeff Wang, Luke Zettlemoyer, Zornitsa Kozareva, Mona Diab, Veselin Stoyanov, and Xian Li. 2022.
\newblock \href {https://doi.org/10.18653/v1/2022.emnlp-main.616} {Few-shot learning with multilingual generative language models}.
\newblock In \emph{Proceedings of the 2022 Conference on Empirical Methods in Natural Language Processing}, pages 9019--9052, Abu Dhabi, United Arab Emirates. Association for Computational Linguistics.

\bibitem[{Liu et~al.(2023)Liu, Cohen, Pasunuru, Choi, Hajishirzi, and Celikyilmaz}]{Liu2023DontTA}
Jiacheng Liu, Andrew Cohen, Ramakanth Pasunuru, Yejin Choi, Hannaneh Hajishirzi, and Asli Celikyilmaz. 2023.
\newblock \href {https://arxiv.org/abs/2309.15028} {Don't throw away your value model! making {PPO} even better via value-guided {M}onte-{C}arlo tree search decoding}.
\newblock \emph{arXiv preprint arXiv:2309.15028}.

\bibitem[{Lu et~al.(2022)Lu, Welleck, West, Jiang, Kasai, Khashabi, Le~Bras, Qin, Yu, Zellers, Smith, and Choi}]{lu-etal-2022-neurologic}
Ximing Lu, Sean Welleck, Peter West, Liwei Jiang, Jungo Kasai, Daniel Khashabi, Ronan Le~Bras, Lianhui Qin, Youngjae Yu, Rowan Zellers, Noah~A. Smith, and Yejin Choi. 2022.
\newblock \href {https://doi.org/10.18653/v1/2022.naacl-main.57} {{N}euro{L}ogic a*esque decoding: Constrained text generation with lookahead heuristics}.
\newblock In \emph{Proceedings of the 2022 Conference of the North American Chapter of the Association for Computational Linguistics: Human Language Technologies}, pages 780--799, Seattle, United States. Association for Computational Linguistics.

\bibitem[{Mathur et~al.(2020)Mathur, Baldwin, and Cohn}]{mathur-etal-2020-tangled}
Nitika Mathur, Timothy Baldwin, and Trevor Cohn. 2020.
\newblock \href {https://doi.org/10.18653/v1/2020.acl-main.448} {Tangled up in {BLEU}: Reevaluating the evaluation of automatic machine translation evaluation metrics}.
\newblock In \emph{Proceedings of the 58th Annual Meeting of the Association for Computational Linguistics}, pages 4984--4997, Online. Association for Computational Linguistics.

\bibitem[{Ng et~al.(2019)Ng, Yee, Baevski, Ott, Auli, and Edunov}]{ng-etal-2019-facebook}
Nathan Ng, Kyra Yee, Alexei Baevski, Myle Ott, Michael Auli, and Sergey Edunov. 2019.
\newblock \href {https://doi.org/10.18653/v1/W19-5333} {{F}acebook {FAIR}{'}s {WMT}19 news translation task submission}.
\newblock In \emph{Proceedings of the Fourth Conference on Machine Translation (Volume 2: Shared Task Papers, Day 1)}, pages 314--319, Florence, Italy. Association for Computational Linguistics.

\bibitem[{{NLLB Team} et~al.(2022){NLLB Team}, Costa-jussà, Cross, Çelebi, Elbayad, Heafield, Heffernan, Kalbassi, Lam, Licht, Maillard, Sun, Wang, Wenzek, Youngblood, Akula, Barrault, Gonzalez, Hansanti, Hoffman, Jarrett, Sadagopan, Rowe, Spruit, Tran, Andrews, Ayan, Bhosale, Edunov, Fan, Gao, Goswami, Guzmán, Koehn, Mourachko, Ropers, Saleem, Schwenk, and Wang}]{nllbteam2022language}
{NLLB Team}, Marta~R. Costa-jussà, James Cross, Onur Çelebi, Maha Elbayad, Kenneth Heafield, Kevin Heffernan, Elahe Kalbassi, Janice Lam, Daniel Licht, Jean Maillard, Anna Sun, Skyler Wang, Guillaume Wenzek, Al~Youngblood, Bapi Akula, Loic Barrault, Gabriel~Mejia Gonzalez, Prangthip Hansanti, John Hoffman, Semarley Jarrett, Kaushik~Ram Sadagopan, Dirk Rowe, Shannon Spruit, Chau Tran, Pierre Andrews, Necip~Fazil Ayan, Shruti Bhosale, Sergey Edunov, Angela Fan, Cynthia Gao, Vedanuj Goswami, Francisco Guzmán, Philipp Koehn, Alexandre Mourachko, Christophe Ropers, Safiyyah Saleem, Holger Schwenk, and Jeff Wang. 2022.
\newblock \href {https://arxiv.org/abs/2207.04672} {No {L}anguage {L}eft {B}ehind: Scaling human-centered machine translation}.
\newblock \emph{arXiv preprint arXiv:2207.04672}.

\bibitem[{Ott et~al.(2018)Ott, Auli, Grangier, and Ranzato}]{Ott2018AnalyzingUI}
Myle Ott, Michael Auli, David Grangier, and Marc'Aurelio Ranzato. 2018.
\newblock \href {https://proceedings.mlr.press/v80/ott18a.html} {Analyzing uncertainty in neural machine translation}.
\newblock In \emph{Proceedings of the 35th International Conference on Machine Learning}, pages 3956--3965.

\bibitem[{Ouyang et~al.(2022)Ouyang, Wu, Jiang, Almeida, Wainwright, Mishkin, Zhang, Agarwal, Slama, Gray, Schulman, Hilton, Kelton, Miller, Simens, Askell, Welinder, Christiano, Leike, and Lowe}]{ouyang2022training}
Long Ouyang, Jeffrey Wu, Xu~Jiang, Diogo Almeida, Carroll Wainwright, Pamela Mishkin, Chong Zhang, Sandhini Agarwal, Katarina Slama, Alex Gray, John Schulman, Jacob Hilton, Fraser Kelton, Luke Miller, Maddie Simens, Amanda Askell, Peter Welinder, Paul Christiano, Jan Leike, and Ryan Lowe. 2022.
\newblock \href {https://openreview.net/forum?id=TG8KACxEON} {Training language models to follow instructions with human feedback}.
\newblock In \emph{Advances in Neural Information Processing Systems}.

\bibitem[{Papineni et~al.(2002)Papineni, Roukos, Ward, and Zhu}]{papineni-etal-2002-bleu}
Kishore Papineni, Salim Roukos, Todd Ward, and Wei-Jing Zhu. 2002.
\newblock \href {https://doi.org/10.3115/1073083.1073135} {{B}leu: a method for automatic evaluation of machine translation}.
\newblock In \emph{Proceedings of the 40th Annual Meeting of the Association for Computational Linguistics}, pages 311--318, Philadelphia, Pennsylvania, USA. Association for Computational Linguistics.

\bibitem[{Popovi{\'c}(2015)}]{popovic-2015-chrf}
Maja Popovi{\'c}. 2015.
\newblock \href {https://doi.org/10.18653/v1/W15-3049} {chr{F}: character n-gram {F}-score for automatic {MT} evaluation}.
\newblock In \emph{Proceedings of the Tenth Workshop on Statistical Machine Translation}, pages 392--395, Lisbon, Portugal. Association for Computational Linguistics.

\bibitem[{Post(2018)}]{post-2018-call}
Matt Post. 2018.
\newblock \href {https://doi.org/10.18653/v1/W18-6319} {A call for clarity in reporting {BLEU} scores}.
\newblock In \emph{Proceedings of the Third Conference on Machine Translation: Research Papers}, pages 186--191, Brussels, Belgium. Association for Computational Linguistics.

\bibitem[{Raffel et~al.(2020)Raffel, Shazeer, Roberts, Lee, Narang, Matena, Zhou, Li, and Liu}]{t5}
Colin Raffel, Noam Shazeer, Adam Roberts, Katherine Lee, Sharan Narang, Michael Matena, Yanqi Zhou, Wei Li, and Peter~J. Liu. 2020.
\newblock Exploring the limits of transfer learning with a unified text-to-text transformer.
\newblock \emph{J. Mach. Learn. Res.}, 21(1).

\bibitem[{Ranasinghe et~al.(2020)Ranasinghe, Orasan, and Mitkov}]{ranasinghe-etal-2020-transquest}
Tharindu Ranasinghe, Constantin Orasan, and Ruslan Mitkov. 2020.
\newblock \href {https://doi.org/10.18653/v1/2020.coling-main.445} {{T}rans{Q}uest: Translation quality estimation with cross-lingual transformers}.
\newblock In \emph{Proceedings of the 28th International Conference on Computational Linguistics}, pages 5070--5081, Barcelona, Spain (Online). International Committee on Computational Linguistics.

\bibitem[{Rei et~al.(2022{\natexlab{a}})Rei, C.~de Souza, Alves, Zerva, Farinha, Glushkova, Lavie, Coheur, and Martins}]{rei-etal-2022-comet}
Ricardo Rei, Jos{\'e}~G. C.~de Souza, Duarte Alves, Chrysoula Zerva, Ana~C Farinha, Taisiya Glushkova, Alon Lavie, Luisa Coheur, and Andr{\'e} F.~T. Martins. 2022{\natexlab{a}}.
\newblock \href {https://aclanthology.org/2022.wmt-1.52} {{COMET}-22: Unbabel-{IST} 2022 submission for the metrics shared task}.
\newblock In \emph{Proceedings of the Seventh Conference on Machine Translation (WMT)}, pages 578--585, Abu Dhabi, United Arab Emirates (Hybrid). Association for Computational Linguistics.

\bibitem[{Rei et~al.(2021)Rei, Farinha, Zerva, van Stigt, Stewart, Ramos, Glushkova, Martins, and Lavie}]{rei-etal-2021-references}
Ricardo Rei, Ana~C Farinha, Chrysoula Zerva, Daan van Stigt, Craig Stewart, Pedro Ramos, Taisiya Glushkova, Andr{\'e} F.~T. Martins, and Alon Lavie. 2021.
\newblock \href {https://aclanthology.org/2021.wmt-1.111} {Are references really needed? unbabel-{IST} 2021 submission for the metrics shared task}.
\newblock In \emph{Proceedings of the Sixth Conference on Machine Translation}, pages 1030--1040, Online. Association for Computational Linguistics.

\bibitem[{Rei et~al.(2020)Rei, Stewart, Farinha, and Lavie}]{rei-etal-2020-comet}
Ricardo Rei, Craig Stewart, Ana~C Farinha, and Alon Lavie. 2020.
\newblock \href {https://doi.org/10.18653/v1/2020.emnlp-main.213} {{COMET}: A neural framework for {MT} evaluation}.
\newblock In \emph{Proceedings of the 2020 Conference on Empirical Methods in Natural Language Processing (EMNLP)}, pages 2685--2702, Online. Association for Computational Linguistics.

\bibitem[{Rei et~al.(2022{\natexlab{b}})Rei, Treviso, Guerreiro, Zerva, Farinha, Maroti, C.~de Souza, Glushkova, Alves, Coheur, Lavie, and Martins}]{rei-etal-2022-cometkiwi}
Ricardo Rei, Marcos Treviso, Nuno~M. Guerreiro, Chrysoula Zerva, Ana~C Farinha, Christine Maroti, Jos{\'e}~G. C.~de Souza, Taisiya Glushkova, Duarte Alves, Luisa Coheur, Alon Lavie, and Andr{\'e} F.~T. Martins. 2022{\natexlab{b}}.
\newblock \href {https://aclanthology.org/2022.wmt-1.60} {{C}omet{K}iwi: {IST}-unbabel 2022 submission for the quality estimation shared task}.
\newblock In \emph{Proceedings of the Seventh Conference on Machine Translation (WMT)}, pages 634--645, Abu Dhabi, United Arab Emirates (Hybrid). Association for Computational Linguistics.

\bibitem[{Sellam et~al.(2020)Sellam, Das, and Parikh}]{sellam-etal-2020-bleurt}
Thibault Sellam, Dipanjan Das, and Ankur Parikh. 2020.
\newblock \href {https://doi.org/10.18653/v1/2020.acl-main.704} {{BLEURT}: Learning robust metrics for text generation}.
\newblock In \emph{Proceedings of the 58th Annual Meeting of the Association for Computational Linguistics}, pages 7881--7892, Online. Association for Computational Linguistics.

\bibitem[{Stahlberg and Byrne(2019)}]{stahlberg-byrne-2019-nmt}
Felix Stahlberg and Bill Byrne. 2019.
\newblock \href {https://doi.org/10.18653/v1/D19-1331} {On {NMT} search errors and model errors: Cat got your tongue?}
\newblock In \emph{Proceedings of the 2019 Conference on Empirical Methods in Natural Language Processing and the 9th International Joint Conference on Natural Language Processing (EMNLP-IJCNLP)}, pages 3356--3362, Hong Kong, China. Association for Computational Linguistics.

\bibitem[{Thompson and Post(2020)}]{thompson-post-2020-automatic}
Brian Thompson and Matt Post. 2020.
\newblock \href {https://doi.org/10.18653/v1/2020.emnlp-main.8} {Automatic machine translation evaluation in many languages via zero-shot paraphrasing}.
\newblock In \emph{Proceedings of the 2020 Conference on Empirical Methods in Natural Language Processing (EMNLP)}, pages 90--121, Online. Association for Computational Linguistics.

\bibitem[{Tomani et~al.(2023)Tomani, Vilar, Freitag, Cherry, Naskar, Finkelstein, and Cremers}]{tomani2023quality}
Christian Tomani, David Vilar, Markus Freitag, Colin Cherry, Subhajit Naskar, Mara Finkelstein, and Daniel Cremers. 2023.
\newblock \href {https://arxiv.org/abs/2310.06707} {Quality control at your fingertips: Quality-aware translation models}.
\newblock \emph{arXiv preprint arXiv:2310.06707}.

\bibitem[{Touvron et~al.(2023)Touvron, Martin, Stone, Albert, Almahairi, Babaei, Bashlykov, Batra, Bhargava, Bhosale, Bikel, Blecher, Ferrer, Chen, Cucurull, Esiobu, Fernandes, Fu, Fu, Fuller, Gao, Goswami, Goyal, Hartshorn, Hosseini, Hou, Inan, Kardas, Kerkez, Khabsa, Kloumann, Korenev, Koura, Lachaux, Lavril, Lee, Liskovich, Lu, Mao, Martinet, Mihaylov, Mishra, Molybog, Nie, Poulton, Reizenstein, Rungta, Saladi, Schelten, Silva, Smith, Subramanian, Tan, Tang, Taylor, Williams, Kuan, Xu, Yan, Zarov, Zhang, Fan, Kambadur, Narang, Rodriguez, Stojnic, Edunov, and Scialom}]{touvron2023llama}
Hugo Touvron, Louis Martin, Kevin Stone, Peter Albert, Amjad Almahairi, Yasmine Babaei, Nikolay Bashlykov, Soumya Batra, Prajjwal Bhargava, Shruti Bhosale, Dan Bikel, Lukas Blecher, Cristian~Canton Ferrer, Moya Chen, Guillem Cucurull, David Esiobu, Jude Fernandes, Jeremy Fu, Wenyin Fu, Brian Fuller, Cynthia Gao, Vedanuj Goswami, Naman Goyal, Anthony Hartshorn, Saghar Hosseini, Rui Hou, Hakan Inan, Marcin Kardas, Viktor Kerkez, Madian Khabsa, Isabel Kloumann, Artem Korenev, Punit~Singh Koura, Marie-Anne Lachaux, Thibaut Lavril, Jenya Lee, Diana Liskovich, Yinghai Lu, Yuning Mao, Xavier Martinet, Todor Mihaylov, Pushkar Mishra, Igor Molybog, Yixin Nie, Andrew Poulton, Jeremy Reizenstein, Rashi Rungta, Kalyan Saladi, Alan Schelten, Ruan Silva, Eric~Michael Smith, Ranjan Subramanian, Xiaoqing~Ellen Tan, Binh Tang, Ross Taylor, Adina Williams, Jian~Xiang Kuan, Puxin Xu, Zheng Yan, Iliyan Zarov, Yuchen Zhang, Angela Fan, Melanie Kambadur, Sharan Narang, Aurelien Rodriguez, Robert Stojnic, Sergey Edunov, and Thomas
  Scialom. 2023.
\newblock \href {https://arxiv.org/abs/2307.09288} {Llama 2: Open foundation and fine-tuned chat models}.
\newblock \emph{arXiv preprint arXiv:2307.09288}.

\bibitem[{Vernikos et~al.(2024)Vernikos, Brazinskas, Adamek, Mallinson, Severyn, and Malmi}]{vernikos-etal-2024-small}
Giorgos Vernikos, Arthur Brazinskas, Jakub Adamek, Jonathan Mallinson, Aliaksei Severyn, and Eric Malmi. 2024.
\newblock \href {https://aclanthology.org/2024.eacl-long.165} {Small language models improve giants by rewriting their outputs}.
\newblock In \emph{Proceedings of the 18th Conference of the European Chapter of the Association for Computational Linguistics (Volume 1: Long Papers)}, pages 2703--2718.

\bibitem[{Vernikos et~al.(2022)Vernikos, Thompson, Mathur, and Federico}]{vernikos-etal-2022-embarrassingly}
Giorgos Vernikos, Brian Thompson, Prashant Mathur, and Marcello Federico. 2022.
\newblock \href {https://aclanthology.org/2022.wmt-1.6} {Embarrassingly easy document-level {MT} metrics: How to convert any pretrained metric into a document-level metric}.
\newblock In \emph{Proceedings of the Seventh Conference on Machine Translation (WMT)}, pages 118--128, Abu Dhabi, United Arab Emirates (Hybrid). Association for Computational Linguistics.

\bibitem[{Vijayakumar et~al.(2018)Vijayakumar, Cogswell, Selvaraju, Sun, Lee, Crandall, and Batra}]{vijayakumar2017diverse}
Ashwin Vijayakumar, Michael Cogswell, Ramprasaath Selvaraju, Qing Sun, Stefan Lee, David Crandall, and Dhruv Batra. 2018.
\newblock \href {https://doi.org/10.1609/aaai.v32i1.12340} {Diverse beam search for improved description of complex scenes}.
\newblock In \emph{Proceedings of the AAAI Conference on Artificial Intelligence}, volume~32.

\bibitem[{Voita et~al.(2019)Voita, Sennrich, and Titov}]{voita-etal-2019-context}
Elena Voita, Rico Sennrich, and Ivan Titov. 2019.
\newblock \href {https://doi.org/10.18653/v1/D19-1081} {Context-aware monolingual repair for neural machine translation}.
\newblock In \emph{Proceedings of the 2019 Conference on Empirical Methods in Natural Language Processing and the 9th International Joint Conference on Natural Language Processing (EMNLP-IJCNLP)}, pages 877--886, Hong Kong, China. Association for Computational Linguistics.

\bibitem[{Wei et~al.(2023)Wei, Wei, Lin, Li, Zhang, Ren, Li, Wan, Cao, Xie, Hu, Li, Hui, Yu, Liu, Yang, Huang, and Xie}]{wei2023polylm}
Xiangpeng Wei, Haoran Wei, Huan Lin, Tianhao Li, Pei Zhang, Xingzhang Ren, Mei Li, Yu~Wan, Zhiwei Cao, Binbin Xie, Tianxiang Hu, Shangjie Li, Binyuan Hui, Bowen Yu, Dayiheng Liu, Baosong Yang, Fei Huang, and Jun Xie. 2023.
\newblock \href {https://arxiv.org/abs/2307.06018} {{PolyLM}: An open source polyglot large language model}.
\newblock \emph{arXiv preprint arXiv:2307.06018}.

\bibitem[{Welleck et~al.(2023)Welleck, Lu, West, Brahman, Shen, Khashabi, and Choi}]{welleck2023generating}
Sean Welleck, Ximing Lu, Peter West, Faeze Brahman, Tianxiao Shen, Daniel Khashabi, and Yejin Choi. 2023.
\newblock \href {https://openreview.net/forum?id=hH36JeQZDaO} {Generating sequences by learning to self-correct}.
\newblock In \emph{The Eleventh International Conference on Learning Representations}.

\bibitem[{Xu et~al.(2024)Xu, Kim, Sharaf, and Awadalla}]{xu2023paradigm}
Haoran Xu, Young~Jin Kim, Amr Sharaf, and Hany~Hassan Awadalla. 2024.
\newblock \href {https://openreview.net/forum?id=farT6XXntP} {A paradigm shift in machine translation: Boosting translation performance of large language models}.
\newblock In \emph{The Twelfth International Conference on Learning Representations}.

\bibitem[{Yee et~al.(2019)Yee, Dauphin, and Auli}]{yee-etal-2019-simple}
Kyra Yee, Yann Dauphin, and Michael Auli. 2019.
\newblock \href {https://doi.org/10.18653/v1/D19-1571} {Simple and effective noisy channel modeling for neural machine translation}.
\newblock In \emph{Proceedings of the 2019 Conference on Empirical Methods in Natural Language Processing and the 9th International Joint Conference on Natural Language Processing (EMNLP-IJCNLP)}, pages 5696--5701, Hong Kong, China. Association for Computational Linguistics.

\bibitem[{Zerva et~al.(2022)Zerva, Blain, Rei, Lertvittayakumjorn, C.~de Souza, Eger, Kanojia, Alves, Or{\u{a}}san, Fomicheva, Martins, and Specia}]{zerva-etal-2022-findings}
Chrysoula Zerva, Fr{\'e}d{\'e}ric Blain, Ricardo Rei, Piyawat Lertvittayakumjorn, Jos{\'e}~G. C.~de Souza, Steffen Eger, Diptesh Kanojia, Duarte Alves, Constantin Or{\u{a}}san, Marina Fomicheva, Andr{\'e} F.~T. Martins, and Lucia Specia. 2022.
\newblock \href {https://aclanthology.org/2022.wmt-1.3} {Findings of the {WMT} 2022 shared task on quality estimation}.
\newblock In \emph{Proceedings of the Seventh Conference on Machine Translation (WMT)}, pages 69--99, Abu Dhabi, United Arab Emirates (Hybrid). Association for Computational Linguistics.

\bibitem[{Zhang et~al.(2020)Zhang, Kishore, Wu, Weinberger, and Artzi}]{Zhang*2020BERTScore:}
Tianyi Zhang, Varsha Kishore, Felix Wu, Kilian~Q. Weinberger, and Yoav Artzi. 2020.
\newblock \href {https://openreview.net/forum?id=SkeHuCVFDr} {{BERTS}core: Evaluating text generation with {BERT}}.
\newblock In \emph{The Eighth International Conference on Learning Representations}.

\bibitem[{Zhu et~al.(2023)Zhu, Liu, Dong, Xu, Huang, Kong, Chen, and Li}]{zhu2023multilingual}
Wenhao Zhu, Hongyi Liu, Qingxiu Dong, Jingjing Xu, Shujian Huang, Lingpeng Kong, Jiajun Chen, and Lei Li. 2023.
\newblock \href {https://arxiv.org/abs/2304.04675} {Multilingual machine translation with large language models: Empirical results and analysis}.
\newblock \emph{arXin preprint arXiv:2304.04675}.

\end{thebibliography}


\appendix

\clearpage 

\section{Appendix}
\label{sec:appendix}

\subsection{Implementation Optimizations} \label{app:optimization}
While Algorithm~\ref{alg:combine_hyps} outlines the concept of fusing candidates, we introduce specific modifications to enhance efficiency. Firstly, to mitigate the computationally expensive calls to the QE model, we parallelize the exploration of all sentences in the test set, resembling a batched beam search. By doing so, we reduce the overall number of calls to the model (which depends on the number of divergent spans) by utilizing a larger batch. Additionally, we implement a hash table to track previously generated candidates, ensuring that we do not compute scores for the same sentence twice. Lastly, we incorporate an early exit mechanism, removing sentences for which no pending pseudo-generation step exists. These optimizations significantly impact the time complexity of our algorithm, which we empirically demonstrate to scale linearly with the number of candidates in Section~\ref{sec:runtime}.

\subsection{Datasets} \label{app:data}
Table~\ref{tab:datasets} presents information about the datasets used in our study, in terms of size and domain.  More information is available in the synthesis articles from WMT22 \cite{freitag-etal-2022-results} and WMT21 \cite{akhbardeh-etal-2021-findings}.

\begin{table}[h]
    \centering
    \small
    \begin{tabular}{c|c|c|c}
        \textbf{Lang. Pair} & \textbf{Source} & \textbf{Sentences} & \textbf{Domain} \\  \hline
         \multirow{4}{*}{en$\rightarrow$de} & \multirow{4}{*}{WMT22} & \multirow{4}{*}{2,037} & News \\ 
         & & & Conversational\\
         & & & e-Commerce\\
         & & & Social\\ \hline
        \multirow{4}{*}{en$\rightarrow$ru} & \multirow{4}{*}{WMT22} & \multirow{4}{*}{2,037} & News\\    
         & & & Conversational\\
         & & & e-Commerce\\
         & & & Social\\ \hline
         \multirow{4}{*}{zh$\rightarrow$en} & \multirow{4}{*}{WMT22} & \multirow{4}{*}{1,875} & News\\ 
         & & & Conversational\\
         & & & e-Commerce\\
         & & & Social\\ \hline
         \multirow{4}{*}{de$\rightarrow$fr} & \multirow{4}{*}{WMT22} & \multirow{4}{*}{1,984} & News\\   & & & Conversational\\
         & & & e-Commerce\\
         & & & Social\\ \hline
         is$\rightarrow$en & WMT21 & 1,000 & News\\ \hline
    \end{tabular}
    \caption{Test datasets used for evaluation.}
    \label{tab:datasets}
\end{table}


\subsection{Results with Surface-based Metrics}
\label{sec:bleu_results}

We provide the scores of the surface-based BLEU and ChrF metrics for LLMs in Table~\ref{tab:llm_results_bleu}. The overall performance trends align with those of neural-based metrics, indicating that larger models consistently achieve higher scores. Once again, QE-fusion consistently surpasses reranking approaches. However, regarding surface-based metrics, beam search or greedy decoding frequently emerge as the top-performing methods, with our approach securing the second position.
Still, our approach outperforms all other sampling-based methods. 

\begin{table*}[ht]
    \centering
    \small
    \begin{tabular}{lccccccc}
    \hline
        & \multicolumn{6}{c}{\bf LLM} \\
        \textbf{Method} &  \textbf{PolyLM-1.7B} &  \textbf{XGLM-2.9B}  &  \textbf{Llama2-7B} & \textbf{Mistral-7B} & \textbf{ALMA-7B} & \textbf{Tower-7B} \\ \hline
        & \multicolumn{6}{c}{en$\rightarrow$de}\\ 
         Greedy & 17.62 / 46.83 & 17.95 / 46.86 & 22.83 / 51.84 & 24.87 / 53.53 & 26.74 / 55.95 & \textbf{30.21} / 58.89\\
        Beam & 12.74 / 46.06 & \textbf{21.16} / 49.04 & 22.99 / \textbf{53.57} & \textbf{24.98} / \textbf{54.79} & \textbf{28.82} / \textbf{57.23} &  29.75 / \textbf{59.80}\\ \hdashline
        Sample & 16.52 / 45.73 & 13.10 / 40.64 & 20.07 / 49.92 & 22.30 / 51.35 & 23.52 / 53.62 & 28.47 / 57.42\\ 
         MBR-BLEU & 19.11 / 47.85 & 18.40 / 46.88 & 22.26 / 51.26 & 23.78 / 52.84 & 25.97 / 55.21 & 30.35 / 58.77\\
         MBR-COMET & 18.94 / 48.22 & 17.65 / 46.84 & 22.37 / 51.95 & 24.18 / 53.40 & 25.91 / 55.64 & 30.51 / 59.03\\
         QE-reranking & 19.45 / 49.07 & 18.55 / 47.95 & 22.38 / 52.12 & 23.90 / 53.32 & 25.73 / 55.63 & 29.77 / 58.81\\ \hdashline
         QE-fusion & \textbf{20.40} / \textbf{50.27} & 19.39 / \textbf{49.11} & \textbf{23.17} / 52.95 & 24.26 / 54.17 & 25.94 / 56.02 & 29.64 / 58.95\\ \hline
         & \multicolumn{6}{c}{en$\rightarrow$ru}\\
        Greedy & 16.36 / 41.81 & 16.68 / 42.29 & 19.64 / 46.11 & 22.64 / 49.04 & 23.14 / 49.75 & \textbf{28.94} / \textbf{54.76}\\
         Beam & 12.63 / 41.46 & \textbf{19.99} / \textbf{44.68} & \textbf{20.52} / 
        \textbf{48.43} & \textbf{23.14} / \textbf{51.26} & \textbf{25.48} / \textbf{51.54} & 27.81 / 55.87\\ \hdashline
        Sample & 14.32 / 39.80 & 11.18 / 34.09 & 17.51 / 43.97 & 20.14 / 46.25 & 20.33 / 47.35 & 26.45 / 52.81\\ 
        MBR-BLEU & 17.42 / 42.31 & 16.38 / 41.20 & 19.51 / 45.85 & 22.27 / 48.24 & 22.77 / 49.25 & 28.76 / 54.41\\
        MBR-COMET & 16.84 / 42.77 & 15.31 / 40.92 & 19.12 / 46.09 & 21.81 / 48.55 & 22.07 / 49.52 & 28.29 / 54.52\\
        QE-reranking & 17.46 / 43.54 & 17.06 / 43.08 & 19.38 / 46.44 & 21.45 / 48.51 & 22.02 / 49.54 & 27.76 / 54.30\\ \hdashline
        QE-fusion & \textbf{18.32} / \textbf{44.89} & 17.34 / 43.95 & 20.04 / 47.32 & 22.06 / 49.44 & 22.43 / 50.30 & 27.94 / \textbf{54.75}\\ \hline
        & \multicolumn{6}{c}{zh$\rightarrow$en}\\ 
        Greedy & 11.22 / 52.34 & 4.44 / 21.40 & 20.37 / 49.25 & 22.92 / 51.68 & 21.61 / 51.14 & 22.16 / 50.20\\
        Beam & 8.74 / 34.81 & \textbf{13.56} / \textbf{38.59} & \textbf{22.37} / \textbf{51.30} & \textbf{24.14} / \textbf{53.69} & \textbf{22.72} / 51.68 & \textbf{23.72} / 51.19\\ \hdashline
        Sample & 9.70 / 34.07 & 5.60 / 24.80 & 17.89 / 46.73 & 20.53 / 49.64 & 19.03 / 49.03 & 20.09 / 48.54\\ 
        MBR-BLEU & 11.20 / 36.60 & 8.93 / 31.87 & 19.98 / 49.01 & 22.39 / 51.70 & 20.87 / 50.79 & 21.90 / 50.26\\
        MBR-COMET & 10.85 / 36.23 & 8.73 / 32.24 & 19.17 / 48.67 & 21.76 / 51.43 & 20.35 / 50.73 & 21.56 / 50.12\\
        QE-reranking & 11.38 / 37.68 & 11.09 / 36.81 & 19.96 / 49.81 & 22.01 / 52.03 & 20.62 / 51.24 & 21.49 / 50.41\\ \hdashline
        QE-fusion & \textbf{12.46} / \textbf{39.36} & 11.58 / 37.64 & 20.44 / 50.72 & 22.31 / 52.63 & 20.93 / \textbf{51.92} & 22.06 / \textbf{51.33}\\ \hline
        & \multicolumn{6}{c}{de$\rightarrow$fr}\\ 
         Greedy & 11.60 / 33.07 & 17.28 / 40.98 & 23.63 / 48.84 & \textbf{26.57} / 52.02 & 20.79 / 46.09 & \textbf{33.69} / 56.59\\
         Beam & 7.79 / 33.73 & \textbf{18.92} / 42.80 & \textbf{23.72} / \textbf{51.25} & 25.17 / \textbf{53.16} & \textbf{21.12} / 45.88 & 32.77 / \textbf{58.07}\\ \hdashline
         Sample & 9.66 / 31.24 & 13.61 / 36.76 & 20.86 / 46.23 & 23.29 / 49.17 & 18.14 / 43.80 & 31.02 / 54.71\\
         MBR-BLEU & 12.15 / 34.57 & 16.45 / 40.93 & 23.14 / 48.45 & 25.54 / 51.05 & 20.37 / 45.83 & 32.90 / 56.36\\
         MBR-COMET & 11.82 / 34.52 & 16.08 / 41.08 & 22.81 / 48.62 & 25.31 / 51.46 & 19.92 / 45.81 & 32.92 / 56.62\\
         QE-reranking & 12.62 / 35.40 & 17.10 / 42.15 & 22.91 / 49.08 & 25.21 / 51.72 & 19.65 / 45.62 & 32.45 / 56.58\\ \hdashline
         QE-fusion & \textbf{13.18} / \textbf{36.31} & 18.10 / \textbf{43.24} & 23.55 / 50.01 & 25.36 / 52.26 & 20.76 / \textbf{46.90} & 31.97 / 56.77\\ \hline
         & \multicolumn{6}{c}{is$\rightarrow$en}\\ 
        Greedy & -- & -- & 14.68 / 36.57 & \textbf{22.60} / 44.73 & 35.72 / 58.20 & 14.84 / 36.76\\
        Beam & -- & -- & 15.47 / 37.03 & 22.49 / 45.64 & \textbf{37.34} / \textbf{59.26} & 14.48 / 37.02\\ \hdashline
        Sample & -- & -- & 13.70 / 35.03 & 20.14 / 43.08 & 32.20 / 55.48 & 13.54 / 35.61\\
        MBR-BLEU & -- & -- & 14.76 / 36.43 & 21.71 / 43.98 & 34.61 / 57.36 & 14.97 / 36.45\\
        MBR-COMET & -- & -- & 14.12 / 36.22 & 21.29 / 44.15 & 34.59 / 57.71 & 14.38 / 36.75\\
        QE-reranking & -- & -- & 15.00 / 37.36 & 21.95 / 45.31 & 34.37 / 57.69 & 15.25 / 37.56\\ \hdashline
        QE-fusion  & -- & -- & \textbf{15.80} / \textbf{38.44} & \textbf{22.62} / \textbf{46.13} & 35.00 / 58.40 & \textbf{15.67} / \textbf{38.32}\\ \hline
    \end{tabular}
    \caption{Translation performance in terms of BLEU~/ ChrF scores for various methods, language pairs, and sizes of LLMs. 
    Dashed lines separate deterministic decoding from existing sampling-based methods and from our approach.}
    \label{tab:llm_results_bleu}
\end{table*}

The outputs of beam search 
often exhibit predictability, 
while sampling introduces a layer of creativity to translations. Unfortunately, surface-based metrics struggle to account for nuances like synonyms or significant restructuring, leading to potential penalties for such translations (a limitation which has contributed to a decline in their popularity).

Table~\ref{tab:mmt_results_bleu} presents BLEU and ChrF scores for the NMT models. Firstly, we observe that the NMT models perform significantly better than the LLMs in terms of surface-based metrics, which is consistent with previous findings \cite{palm, hendy2023good, zhu2023multilingual}.  The trend differs slightly compared to LLMs, as methods like QE-reranking and MBR score higher in terms of surface-based metrics, particularly for the high-resource pairs en$\rightarrow$de and en$\rightarrow$ru. This divergence can be attributed, once again, to the limited diversity of translations generated by MT models, while our approach favors more creative translations, potentially penalized by these metrics. Nevertheless, for pairs where MT models perform worse, such as is$\rightarrow$en and de$\rightarrow$fr, QE-fusion consistently outperforms reranking approaches and occasionally beam search.

\begin{table}[ht]
    \centering
    \small
    \begin{tabular}{lcc}
        \hline
        & \multicolumn{2}{c}{\bf Multilingual NMT} \\
        \textbf{Method} &  \textbf{NLLB-1.3B} &  \textbf{NLLB-3.3B} \\ \hline
        & \multicolumn{2}{c}{en$\rightarrow$de} \\ 
        Greedy & 32.04 / 59.16 & 33.31 / 60.49\\
        Beam  & \textbf{33.69} / \textbf{60.92} & \textbf{34.07} / \textbf{61.82} \\ \hdashline
        Sample & 29.80 / 57.71 & 30.42 / 58.51\\ 
         MBR-BLEU & 31.49 / 58.48 & 32.19 / 59.81\\
         MBR-COMET & 31.19 / 59.00 & 32.11 / 60.18\\
         QE-reranking & 31.01 / 59.26 & 31.68 / 60.10 \\ \hdashline
         QE-fusion & 30.90 / 59.41 & 31.59 / 60.32\\ \hline
        &  \multicolumn{2}{c}{en$\rightarrow$ru} \\ 
        Greedy & 27.58 / 53.37 & 29.12 / 54.71\\
        Beam & \textbf{29.14} / \textbf{54.86} & \textbf{30.12} / \textbf{55.96}\\ \hdashline
        Sample & 25.86 / 52.14 & 27.22 / 53.16\\
        MBR-BLEU & 27.34 / 53.19 & 28.62 / 54.36\\
        MBR-COMET & 27.42 / 53.51 & 28.55 / 54.50\\
        QE-reranking & 27.18 / 53.55 & 28.20 / 54.46\\ \hdashline
        QE-fusion & 26.98 / 53.60 & 28.18 / 54.74\\ \hline
        & \multicolumn{2}{c}{zh$\rightarrow$en}\\ 
        Greedy & 16.71 / 44.13 & 16.66 / 45.56\\
        Beam & \textbf{18.43} / \textbf{47.05} & \textbf{18.34} / 47.72\\ \hdashline
        Sample & 14.47 / 42.83 & 14.66 / 43.75\\
        MBR-BLEU & 16.52 / 45.26 & 17.30 / 46.57\\
        MBR-COMET & 16.42 / 45.54 & 16.98 / 46.38\\
        QE-reranking & 16.70 / 46.27 & 17.40 / 47.42\\ \hdashline
        QE-fusion & 17.13 / 47.00 & 17.51 / \textbf{47.98}\\ \hline
        & \multicolumn{2}{c}{de$\rightarrow$fr}\\ 
        Greedy & 34.56 / 57.65 & 35.88 / 58.33 \\
        Beam &  \textbf{37.04} / \textbf{59.89} & \textbf{38.68} / \textbf{61.14} \\ \hdashline
        Sample & 30.94 / 55.10 & 31.64 / 55.08 \\
        MBR-BLEU & 33.42 / 56.90 & 34.74 / 57.77 \\
        MBR-COMET & 33.38 / 57.25 & 34.40 / 57.87 \\
        QE-reranking & 33.07 / 57.20 & 34.01 / 57.88 \\ \hdashline
        QE-fusion & 33.13 / 57.62 & 33.54 / 58.12 \\ \hline
        & \multicolumn{2}{c}{is$\rightarrow$en}\\ 
        Greedy & 31.62 / 54.93 & 32.60 / 55.56 \\
        Beam & \textbf{32.96} / \textbf{56.25} & \textbf{34.10} / \textbf{57.01} \\ \hdashline
        Sample & 28.78 / 52.95 & 29.35 / 53.53 \\
        MBR-BLEU & 30.83 / 54.29 & 31.71 / 55.16 \\
        MBR-COMET & 30.25 / 54.17 & 31.56 / 55.33 \\
        QE-reranking & 30.60 / 54.48 & 31.10 / 55.31 \\ \hdashline
        QE-fusion & 31.03 / 54.94 & 31.61 / 55.62 \\ \hline
    \end{tabular}
    \caption{Translation performance in terms of BLEU~/ ChrF scores for various methods, language pairs, and multilingual NMT models.}
    \label{tab:mmt_results_bleu}
\end{table}

\subsection{Results using LLMs with 13B Parameters} 
\label{app:13b}

Tables~\ref{tab:extra_llm_results} and \ref{tab:extra_llm_results_bleu} present the translation scores of the larger versions of Llama2 and ALMA with 13B parameters. The overall trend aligns with other LLMs: QE-fusion significantly outperforms reranking methods.

\begin{table}
    \centering
    \small
    \begin{tabular}{lcc}
    \hline
        & \multicolumn{2}{c}{\bf LLM} \\
        \textbf{Method} & \textbf{Llama2-13B} & \textbf{ALMA-13B} \\ \hline
        & \multicolumn{2}{c}{en$\rightarrow$de}\\ 
        Greedy & 82.88 / 71.62 & 84.66 / 74.08\\ 
        Beam & 84.11 / 73.29 & 85.67 / 75.40\\\hdashline
        Sample & 81.48 / 70.20 & 83.83 / 73.11\\ 
        MBR-BLEU & 82.36 / 71.03 & 84.37 / 73.77\\
        MBR-COMET & 85.08 / 73.41 & 86.04 / 75.40\\
        QE-reranking & 84.61 / 73.96 & 86.03 / 75.75\\ \hdashline
        QE-fusion & \textbf{85.17} / \textbf{74.41} & \textbf{86.32} / \textbf{76.09}\\ \hline
        & \multicolumn{2}{c}{en$\rightarrow$ru}\\ 
        Greedy & 84.73 / 70.68 & 86.93 / 73.67\\
        Beam & 86.14 / 72.71 & 88.05 / 75.51\\ \hdashline
        Sample & 83.42 / 68.83 & 85.97 / 72.48\\ 
        MBR-BLEU & 84.19 / 69.94 & 86.80 / 73.42\\
        MBR-COMET & 87.01 / 72.44 & \textbf{88.54} / 75.12\\
        QE-reranking & 86.60 / 72.96 & 88.17 / 75.42\\ \hdashline
        QE-fusion & \textbf{87.20} / \textbf{73.60} & \textbf{88.53} / \textbf{76.03}\\ \hline
        & \multicolumn{2}{c}{zh$\rightarrow$en}\\ 
        Greedy & 79.37 / 66.27 & 80.04 / 67.18\\
        Beam & 79.67 / 67.30 & 80.51 / 68.15\\ \hdashline
        Sample & 78.53 / 65.19 & 79.28 / 66.05\\ 
        MBR-BLEU & 79.28 / 66.08 & 79.83 / 66.83\\
        MBR-COMET & 80.60 / 66.97 & 81.20 / 67.76\\
        QE-reranking & 80.54 / 67.60 & 81.13 / 68.44\\ \hdashline
        QE-fusion & \textbf{80.94} / \textbf{68.04} & \textbf{81.58} / \textbf{68.96}\\ \hline
        & \multicolumn{2}{c}{de$\rightarrow$fr}\\ 
        Greedy & 78.94 / 63.88 &  77.58 / 60.69\\
        Beam & 80.08 / 65.86 &  78.30 / 61.70\\ \hdashline
        Sample & 77.88 / 61.84 & 75.53 / 57.78\\
        MBR-BLEU & 78.88 / 63.42 & 77.01 / 59.62\\
        MBR-COMET & 81.35 / 65.73 & \textbf{79.88} / 62.39\\
        QE-reranking & 80.99 / 66.43 & 79.24 / 62.52\\ \hdashline
        QE-fusion & \textbf{81.63} / \textbf{67.11} & 79.73 / \textbf{63.36}\\ \hline
        & \multicolumn{2}{c}{is$\rightarrow$en}\\ 
        Greedy & 71.02 / 56.63 & 85.94 / 75.18\\
        Beam & 71.94 / 58.14 & 86.16 / 75.47\\ \hdashline
        Sample & 70.63 / 56.02 & 85.21 / 74.16\\
        MBR-BLEU & 71.18 / 56.71 & 85.96 / 75.04\\
        MBR-COMET & 73.72 / 58.12 & 86.67 / 75.65\\
        QE-reranking & 74.00 / 59.93 & 86.65 / 76.04\\ \hdashline
        QE-fusion & \textbf{74.90} / \textbf{61.17} & \textbf{86.82} / \textbf{76.21}\\ \hline
    \end{tabular}
    \caption{Translation performance of LLMs (13 billion parameters) in terms of COMET~/ BLEURT scores for various methods and language pairs.}
    \label{tab:extra_llm_results}
\end{table}

\begin{table}[ht]
    \centering
    \small
    \begin{tabular}{lccccccc} 
    \hline
        & \multicolumn{2}{c}{\bf LLM} \\
        \textbf{Method} &  \textbf{Llama2-13B} & \textbf{ALMA-13B} \\ \hline
        & \multicolumn{2}{c}{en$\rightarrow$de}\\ 
         Greedy & 26.40 / 54.83 & 28.56 / 57.35\\
         Beam & \textbf{26.43} / \textbf{56.22} & \textbf{30.27} / \textbf{59.04}\\ \hdashline
         Sample & 23.96 / 52.94 & 26.46 / 55.68 \\ 
         MBR-BLEU & 25.39 / 54.04 & 28.53 / 56.88\\
         MBR-COMET & 25.27 / 54.45 & 28.26 / 57.10\\
         QE-reranking & 25.57 / 54.81 & 28.33 / 57.52\\ \hdashline
         QE-fusion & 26.08 / 55.49 & 28.10 / 57.66\\ \hline
         & \multicolumn{2}{c}{en$\rightarrow$ru}\\
        Greedy & 23.30 / 49.84 & 26.56 / 52.12\\
         Beam & \textbf{23.72} / \textbf{51.17} & \textbf{27.86} / \textbf{54.04}\\ \hdashline
        Sample & 21.10 / 47.81 & 23.55 / 49.80\\ 
        MBR-BLEU & 22.76 / 49.14 & 25.49 / 51.38\\
        MBR-COMET & 22.67 / 49.59 & 25.13 / 51.54\\
        QE-reranking & 22.56 / 49.51 & 24.75 / 51.44\\ \hdashline
        QE-fusion & 22.90 / 50.24 & 24.92 / 52.05\\ \hline
        & \multicolumn{2}{c}{zh$\rightarrow$en}\\ 
        Greedy & 22.48 / 51.63 & 24.27 / 53.92\\
        Beam & \textbf{24.48} / \textbf{53.73} & \textbf{26.58} / \textbf{55.13}\\ \hdashline
        Sample & 19.70 / 49.36 & 21.73 / 51.81\\ 
        MBR-BLEU & 21.98 / 51.41 & 23.46 / 53.52\\
        MBR-COMET & 21.42 / 51.34 & 23.05 / 53.59\\
        QE-reranking & 21.51 / 51.63 & 23.37 / 53.93\\ \hdashline
        QE-fusion & 21.70 / 52.02 & 23.32 / 54.36\\ \hline
        & \multicolumn{2}{c}{de$\rightarrow$fr}\\ 
        Greedy & 26.67 / 51.87 & \textbf{23.11} / \textbf{47.54}\\
        Beam & 25.49 / \textbf{53.13} & 21.69 / 45.03\\ \hdashline
        Sample & 24.33 / 49.44 & 20.46 / 44.98\\
        MBR-BLEU & 26.19 / 51.31 & 22.56 / 46.83\\
        MBR-COMET & 25.97 / 51.68 & 22.55 / 47.04\\
        QE-reranking & 26.62 / 52.18 & 21.82 / 46.54\\ \hdashline
        QE-fusion & \textbf{26.71} / 52.70 & 22.39 / 47.35\\ \hline
        & \multicolumn{2}{c}{is$\rightarrow$en}\\ 
        Greedy & 19.82 / 41.59 & 34.75 / 57.55\\
        Beam & 19.92 / 42.40 & \textbf{35.72} / 58.00\\ \hdashline
        Sample & 17.48 / 39.97 & 31.05 / 54.90\\
        MBR-BLEU & 19.50 / 41.43 & 33.52 / 56.70\\
        MBR-COMET & 18.99 / 41.32 & 33.33 / 56.62\\
        QE-reranking & 19.53 / 42.12 & 34.27 / 57.74\\ \hdashline
        QE-fusion  & \textbf{20.40} / \textbf{43.14} & 34.48 / \textbf{58.12}\\ \hline
    \end{tabular}
    \caption{Translation performance of LLMs (13 billion parameters) in terms of BLEU~/ ChrF scores for various methods and language pairs.}
    \label{tab:extra_llm_results_bleu}
\end{table}

\subsection{Graphical Comparison of Main Scores}
\label{sec:radar-chart}

\begin{figure*}
    \centering
    \includegraphics[width=0.9\linewidth]{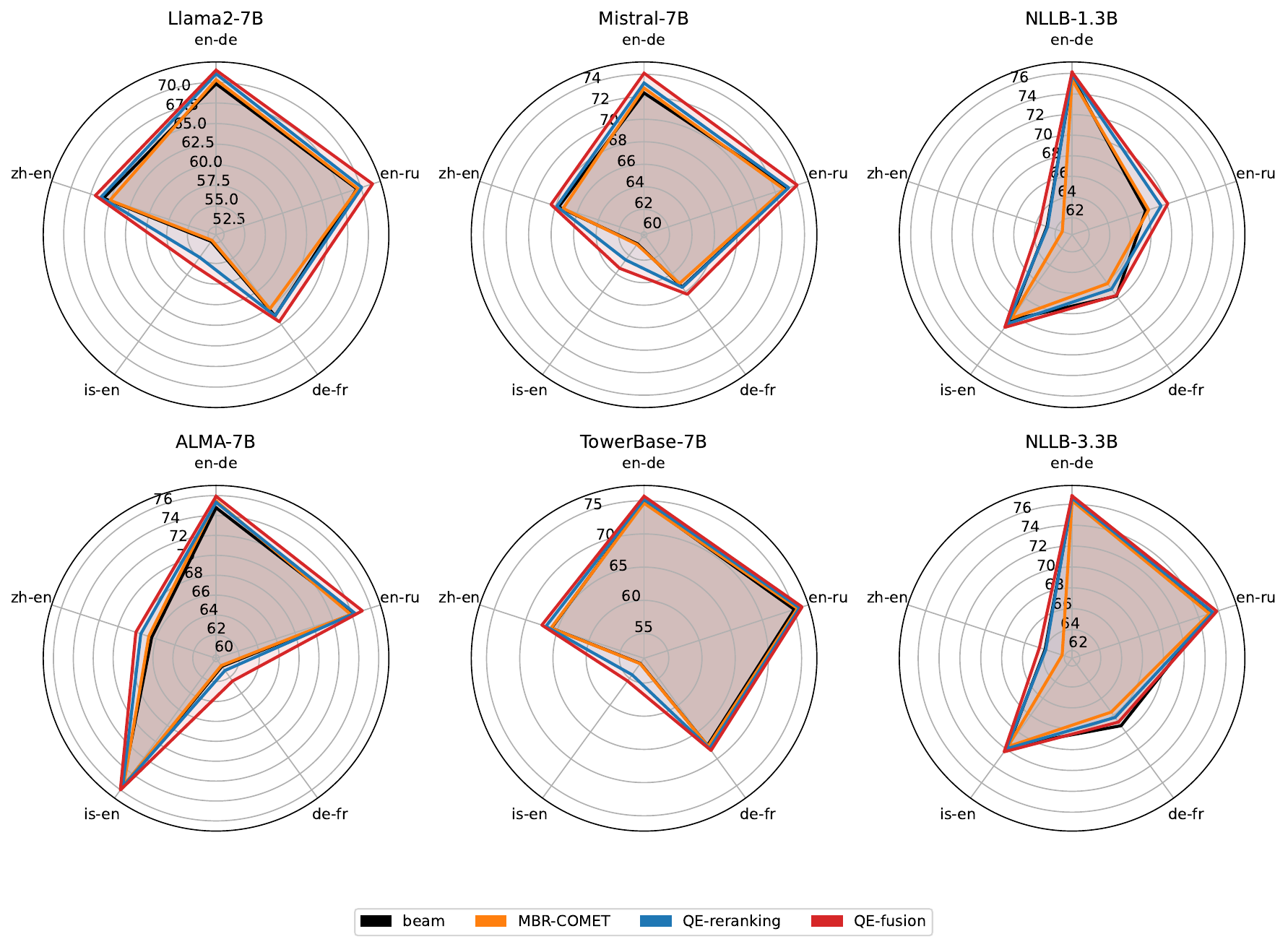}
    \caption{BLEURT scores for four methods combined with four LLMs and two NMT models, on five and four language pairs respectively.}
    \label{fig:spider_joint}
\end{figure*}

The BLEURT scores of various methods, LLMs and language pairs from Table~\ref{tab:llm_results} are represented as radar charts in Figure~\ref{fig:spider_joint}. The shapes corresponding to our proposal, QE-fusion, always extend outside the others (only beam search, MBR with COMET and QE-reranking are plotted, for simplicity), for all of the four LLMs represented: Llama2-7B, Mistral-7B, ALMA-7B, and TowerBase-7B.  The latter two models, though fine-tuned for MT, have large differences for is$\rightarrow$en and de$\rightarrow$fr.

Figure~\ref{fig:spider_joint} also shows the BLEURT scores 
for the two NMT models from Table~\ref{tab:mmt_results}, over the same language pairs as above, excluding  zh$\rightarrow$en.  Again, QE-fusion extends outside the other shapes for both NMT models.

\subsection{Temperature and Diversity}
\label{app:temp-div}

\begin{figure}[ht]
    \centering
    \includegraphics[width=0.85\linewidth]{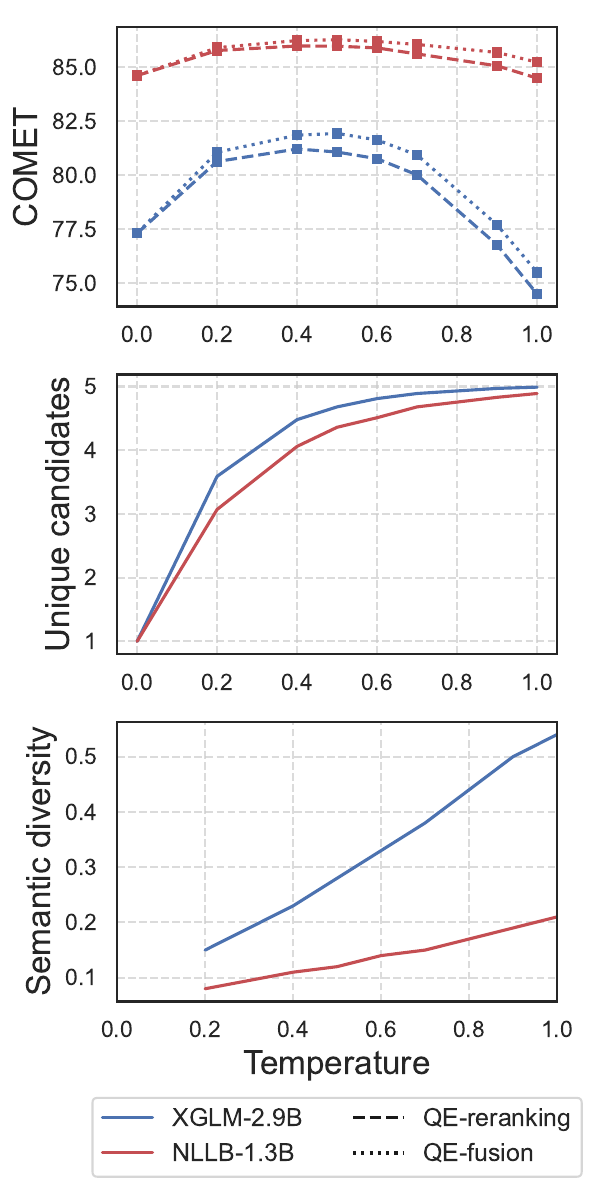}
    \caption{Effect of temperature on the diversity of the pool (below) and its resulting impact on translation performance (above) using LLMs and NMT models for en$\rightarrow$de translation.}
    \label{fig:temp_additional}
\end{figure}

In Figure~\ref{fig:temp_additional}, we present additional results regarding the impact of temperature on translation performance and pool diversity. We evaluate translation quality with COMET, demonstrating similar results to those in Section~\ref{sec:temp} with BLEURT. Higher temperatures enhance the results of both QE-reranking and QE-fusion but excessively high temperatures lead to a drop in performance.

To measure diversity, we consider here two additional metrics: the average number of unique candidates in the pool and the semantic diversity, as defined by \citet{farinhas-etal-2023-empirical}, where $u(x,y)$ is the utility function, in this case COMET, and $y_j$, $y_i$ represent two different candidates from the pool: 
\begin{equation}
1 - \frac{1}{{N(N - 1)}} \sum_{\substack{i,j=1 \\ j \neq i}}^N u(y_j, y_i) 
\end{equation}
The diversity metrics exhibit similar trends to our lexical diversity ones, with diversity increasing as the temperature rises. 
These results confirm that LLMs tend to produce more diverse outputs, a fact that contributes to explaining why QE-fusion is more effective on LLM outputs than on NMT ones. 

\end{document}